\documentclass[10pt,twocolumn,letterpaper]{article}

\usepackage[pagenumbers]{cvpr} 

\usepackage{graphicx}
\usepackage{amsmath}
\usepackage{amssymb}
\usepackage{booktabs}

\usepackage{adjustbox}
\usepackage{pifont}
\usepackage[misc]{ifsym}

\usepackage[accsupp]{axessibility}  

\newif\ifshowcomments
\showcommentstrue

\ifshowcomments

\newcommand{\TODO}[1]{{\color{red}{[TODO: #1]}}}

\newcommand{\revised}[1]{{\color[rgb]{0.2,0.7,0.2}{#1}}}

\else
\newcommand{\TODO}[1]{}
\newcommand{\revised}[1]{}
\fi

\usepackage[pagebackref,breaklinks,colorlinks]{hyperref}
\usepackage{tabularx}

\usepackage[capitalize]{cleveref}
\crefname{section}{Sec.}{Secs.}
\Crefname{section}{Section}{Sections}
\Crefname{table}{Table}{Tables}
\crefname{table}{Tab.}{Tabs.}

\def\cvprPaperID{3725} 
\def\confName{CVPR}
\def\confYear{2023}

\makeatletter
\def\thanks#1{\protected@xdef\@thanks{\@thanks
        \protect\footnotetext{#1}}}
\makeatother

\begin{document}

\title{Video Dehazing via a Multi-Range Temporal Alignment \\ Network with Physical Prior}

\author{Jiaqi Xu \textsuperscript{\rm 1, 2, $^{\star}$},
   Xiaowei Hu \textsuperscript{\rm 2, $^{\textrm{\Letter}}$},
   Lei Zhu \textsuperscript{\rm 3, 4, $^{\textrm{\Letter}}$},
   Qi Dou \textsuperscript{\rm 1},
   Jifeng Dai \textsuperscript{\rm 5, 2},
   Yu Qiao  \textsuperscript{\rm 2},
   Pheng-Ann Heng \textsuperscript{\rm 1} \\
 \textsuperscript{\rm 1} The Chinese University of Hong Kong \quad
 \textsuperscript{\rm 2} Shanghai Artificial Intelligence Laboratory \\
 \textsuperscript{\rm 3} The Hong Kong University of Science and Technology (Guangzhou) \\
 \textsuperscript{\rm 4} The Hong Kong University of Science and Technology \quad
 \textsuperscript{\rm 5} Tsinghua University
}
\thanks{ 
\hspace{-6mm}  ${\star}$ : This work was done during Jiaqi Xu's internship at Shanghai Artificial Intelligence Laboratory. \\
 $^{\textrm{\Letter}}$ : Corresponding authors (huxiaowei@pjlab.org.cn; leizhu@ust.hk).}

\maketitle

\begin{abstract}
   Video dehazing aims to recover haze-free frames with high visibility and contrast.
   This paper presents a novel framework to effectively explore the physical haze priors and aggregate temporal information.
   Specifically, we design a memory-based physical prior guidance module to encode the prior-related features into long-range memory.
   Besides, we formulate a multi-range scene radiance recovery module to capture space-time dependencies in multiple space-time ranges, which helps to effectively aggregate temporal information from adjacent frames.
   Moreover, we construct the first large-scale outdoor video dehazing benchmark dataset, which contains videos in various real-world scenarios.
   Experimental results on both synthetic and real conditions show the superiority of our proposed method.
   \vspace{-3mm}
\end{abstract}

\section{Introduction}
\label{sec:intro}

Haze largely degrades the visibility and contrast of the outdoor scenes, which adversely affects the performance of downstream vision tasks, such as the detection and segmentation in autonomous driving and surveillance.
%
%
According to the atmospheric scattering model~\cite{mccartney1976optics,he2010single}, the formation of a hazy image is described as:
\begin{equation}
    \label{eq:physical model}
    I(x)=J(x)t(x)+A(1-t(x)) \ ,
\end{equation}
where $I, J, A, t$ denote the observed hazy image, scene radiance, atmospheric light, and transmission, respectively, and $x$ is the pixel index.
%
The transmission $t = e^{-\beta d(x)}$ describes the scene radiance attenuation caused by the light scattering, where $\beta$ is the scattering coefficient of the atmosphere, and $d$ denotes the scene depth.
%
%

%
%
%

\begin{figure}
    \centering
    \captionsetup[subfigure]{justification=centering}
    \begin{subfigure}{0.48\hsize}
        \includegraphics[width=\textwidth]{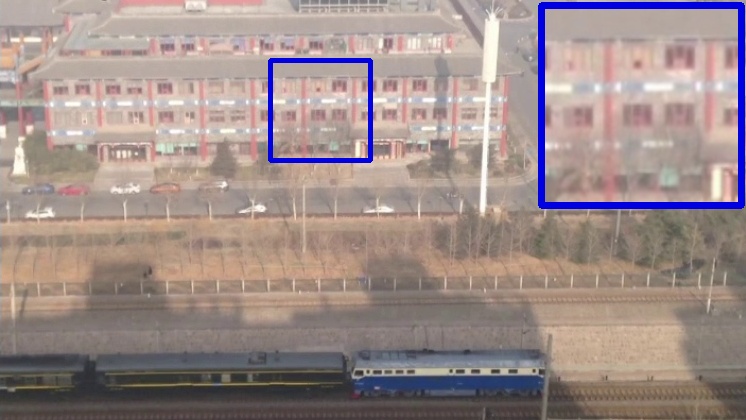}
        \caption{Hazy frame}
    \end{subfigure}
    \begin{subfigure}{0.48\hsize}
        \includegraphics[width=\textwidth]{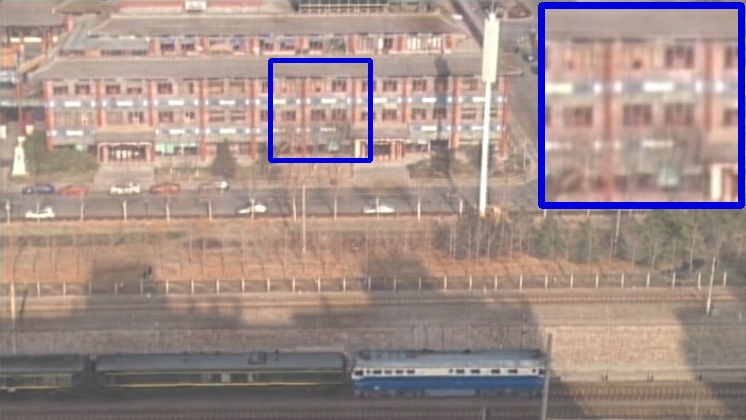}
        \caption{VDH~\cite{ren2018deep}}
    \end{subfigure}
    \\
    \begin{subfigure}{0.48\hsize}
        \includegraphics[width=\textwidth]{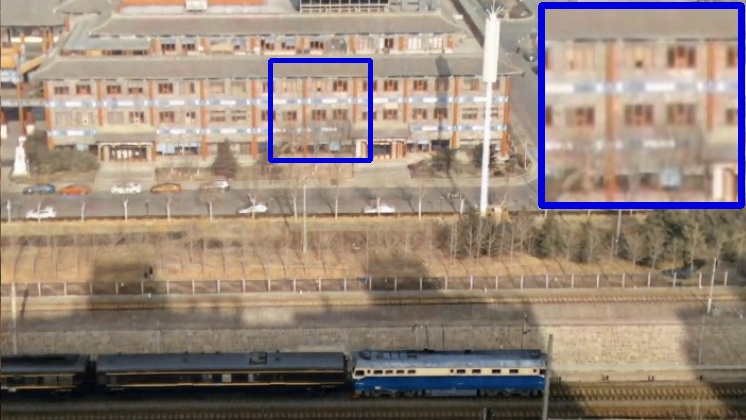}
        \caption{CG-IDN~\cite{zhang2021learning}}
    \end{subfigure}
    \begin{subfigure}{0.48\hsize}
        \includegraphics[width=\textwidth]{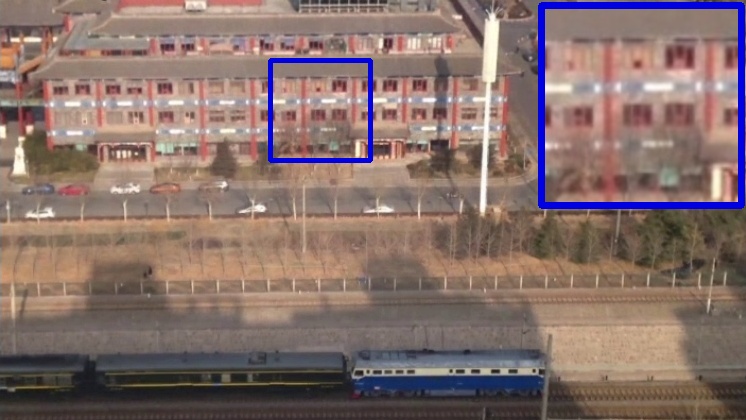}
        \caption{Our method}
    \end{subfigure}
    \\
    \vspace{-2mm}
    \caption{Visual comparison on a real-world hazy video. Our method trained on our outdoor dehazing dataset clearly removes haze without color distortion.}
    \label{fig:comparison}
    \vspace{-6mm}
\end{figure}

Video dehazing benefits from temporal clues, such as highly correlated haze thickness and lighting conditions, as well as the moving foreground objects and backgrounds.
%
%
Early deep learning-based video dehazing methods leverage temporal information by simply concatenating input frames or feature maps~\cite{ren2018deep,li2018end}.
Recently, GC-IDN~\cite{zhang2021learning} proposes to use cost volume and confidence to align and aggregate temporal information.
However, existing video dehazing methods suffer from several limitations.
First, these approaches either obtain haze-free frames from the physical model-based component estimation~\cite{ren2018deep,li2018end} or ignore the explicit physical prior embedded in the haze imaging model~\cite{zhang2021learning}.
The former suffers from inaccurate intermediate prediction, thus leading to error accumulation in the final results, while the latter overlooks the physical prior information, which plays an important role in haze estimation and scene recovery.
Second, these methods aggregate temporal information by using input/feature stacking or frame-to-frame alignment in a local sliding window, which is hard to obtain global and long-range temporal information.
%

In this work, we present a novel video dehazing framework via a Multi-range temporal Alignment network with Physical prior (MAP-Net) to address the aforementioned issues.
%
%
First, we design a memory-based physical prior guidance module, which aims to inject the physical prior to help the scene radiance recovery.
Specifically, we perform feature disentanglement according to the physical model with two decoders, where one estimates the transmission and atmospheric light, and the other recovers scene radiance.
The feature extracted from the first decoder is leveraged as the physical haze prior, which is integrated into the second decoder for scene radiance recovery.
To infer the global physical prior in a long-range video, we design a physical prior token memory that effectively encodes prior-related features into compact tokens for efficient memory reading.
%
%
%

Second, we introduce a multi-range scene radiance recovery module to capture space-time dependencies in multiple space-time ranges.
This module first splits the adjacent frames into multiple ranges, then aligns and aggregates the corresponding recurrent range features, and finally recovers the scene radiance.
%
Unlike CG-IDN~\cite{zhang2021learning}, which aligns the adjacent features frame-by-frame, we align the features of adjacent frames into multiple sets with different ranges, which helps to explore the temporal haze clues in various time intervals.
%
We further design a space-time deformable attention to warp the features of multiple ranges to the target frame, followed by a guided multi-range complementary information aggregation.
Also, we use an unsupervised flow loss to encourage the network to focus on the aligned areas and train the whole network in an end-to-end manner.

In addition, the existing learning-based video dehazing methods are mainly trained and evaluated on indoor datasets~\cite{ren2018deep,li2018end,zhang2021learning}, which suffer from performance degradation in real-world outdoor scenarios.
%
%
Thus, we construct an outdoor video dehazing benchmark dataset, HazeWorld, which has three main properties.
First, it is a large-scale synthetic dataset with 3,588 training videos and 1,496 testing videos.
Second, we collect videos from diverse outdoor scenarios, \eg, autonomous driving and life scenes.
Third, the dataset has various downstream tasks for evaluation, such as segmentation and detection.
%
%
Various experiments on both synthetic and real datasets demonstrate the effectiveness of our approach, which clearly outperforms the existing image and video dehazing methods; see Fig.~\ref{fig:comparison}.
%
The code and dataset are publicly available at \url{https://github.com/jiaqixuac/MAP-Net}.

Our main contributions are summarized as follows:
%
\begin{itemize}
\itemsep0em 
  \item We present a novel framework, MAP-Net, for video dehazing. A memory-based physical prior guidance module is designed to enhance the scene radiance recovery, which encodes haze-prior-related features into long-range memory.
  \item We introduce a recurrent multi-range scene radiance recovery module with the space-time deformable attention and the guided multi-range aggregation, which effectively captures long-range temporal haze and scene clues from the adjacent frames.
  \item We construct a large-scale outdoor video dehazing dataset with diverse real-world scenarios and labels for downstream task evaluation.
  \item Extensive experiments on both synthetic and real conditions demonstrate our superior performance against the recent state-of-the-art methods.
\end{itemize}

\section{Related Work}

\begin{figure*}
    \centering
    \captionsetup[subfigure]{justification=centering}
    \begin{subfigure}{0.147\textwidth}
        \includegraphics[width=\textwidth]{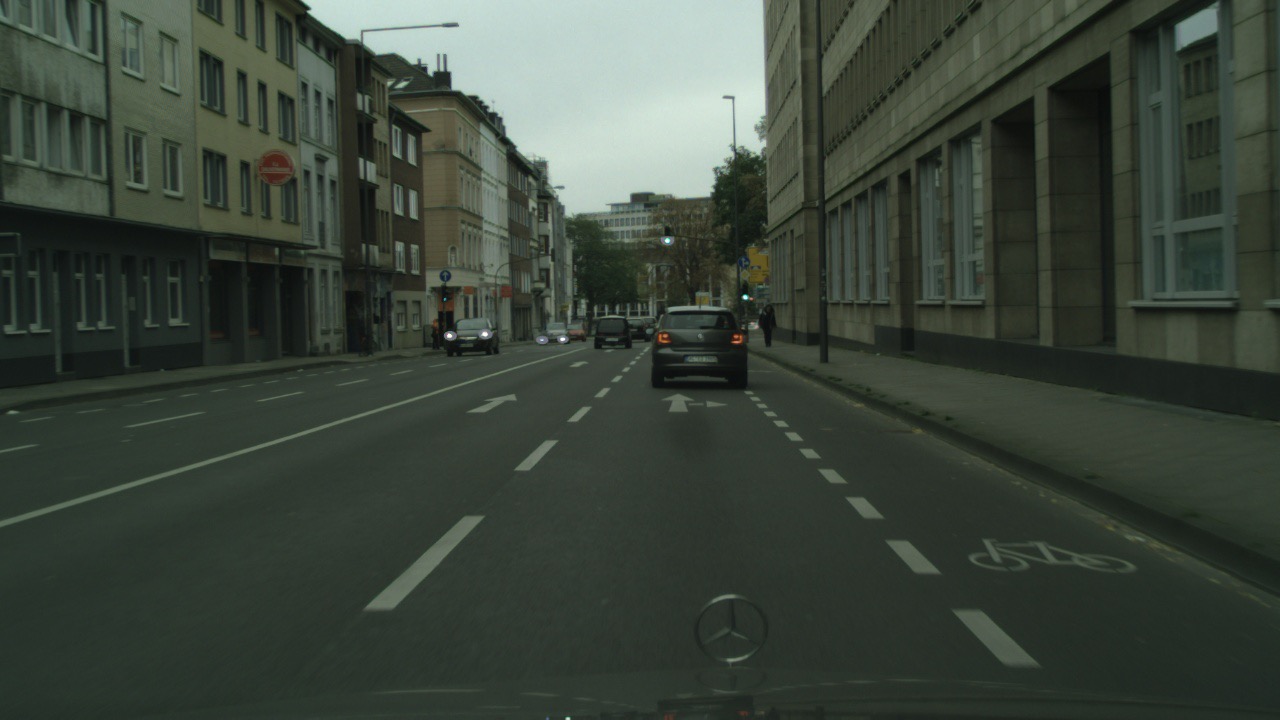}
    \end{subfigure}
    \begin{subfigure}{0.147\textwidth}
        \includegraphics[width=\textwidth]{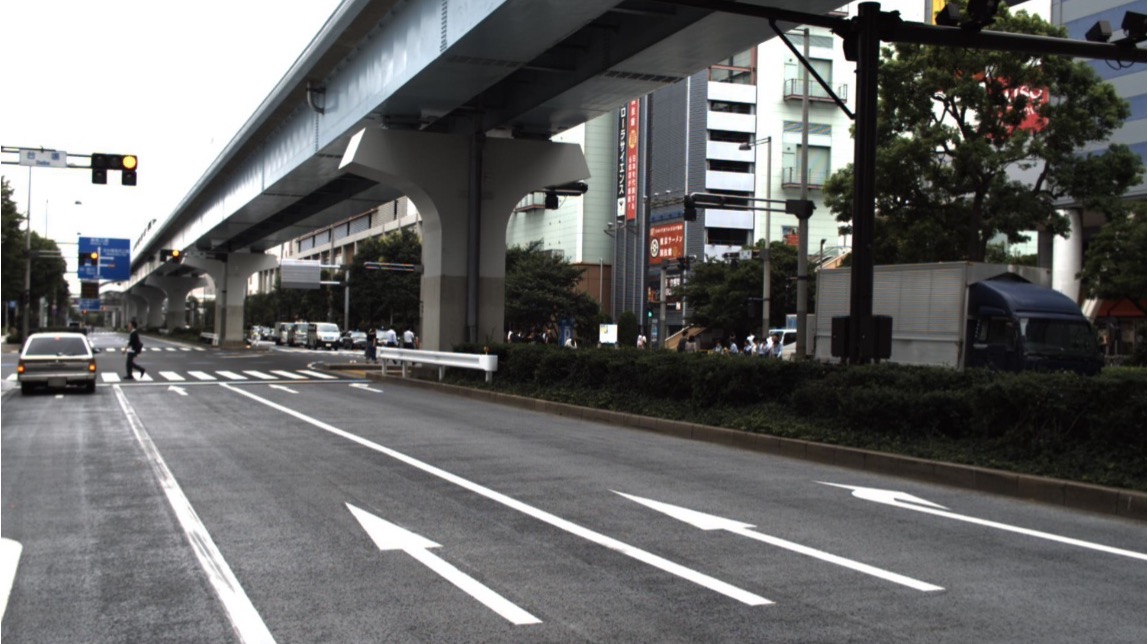}
    \end{subfigure}
    \begin{subfigure}{0.147\textwidth}
        \includegraphics[width=\textwidth]{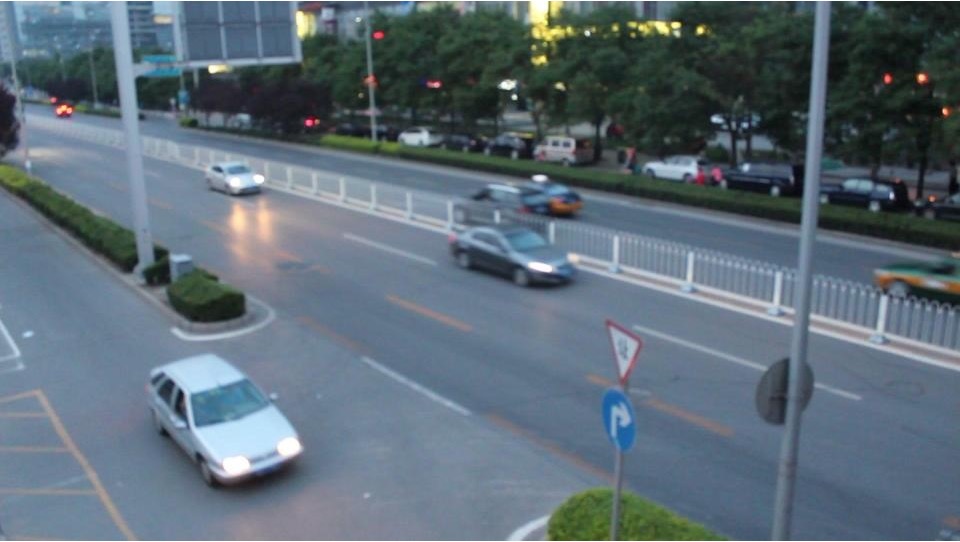}
    \end{subfigure}
    \begin{subfigure}{0.147\textwidth}
        \includegraphics[width=\textwidth]{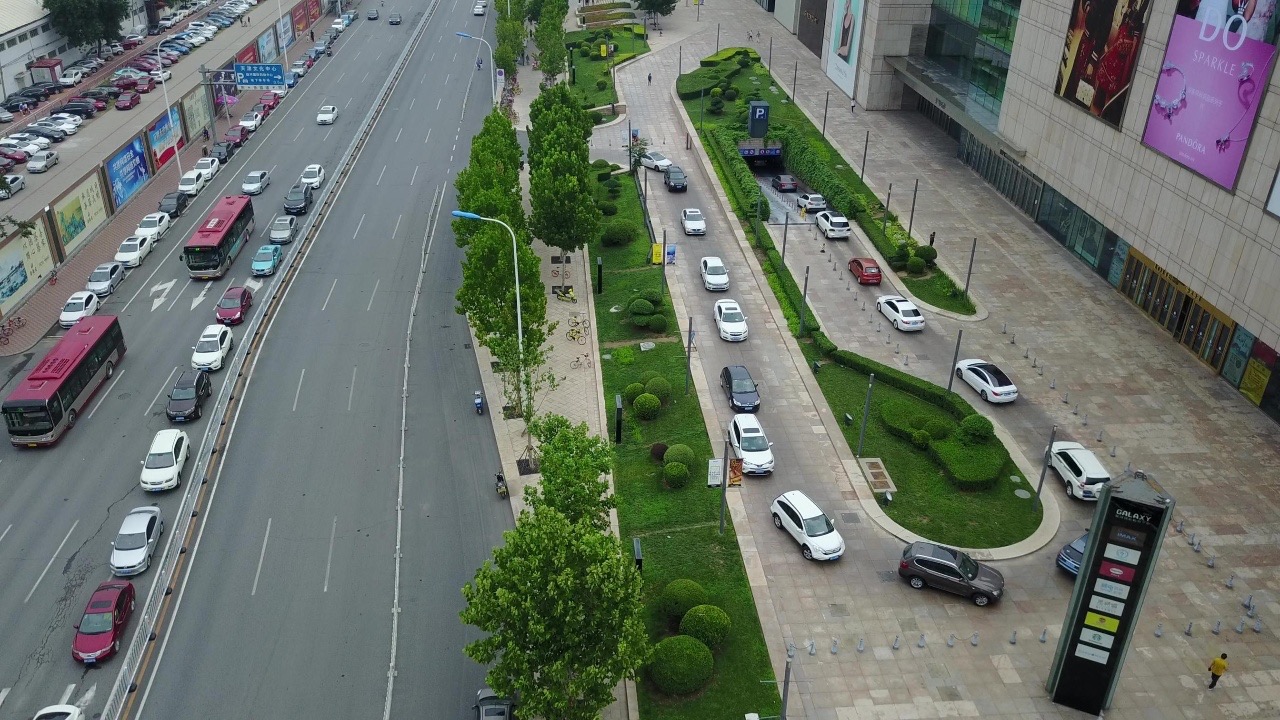}
    \end{subfigure}
    \begin{subfigure}{0.147\textwidth}
        \includegraphics[width=\textwidth]{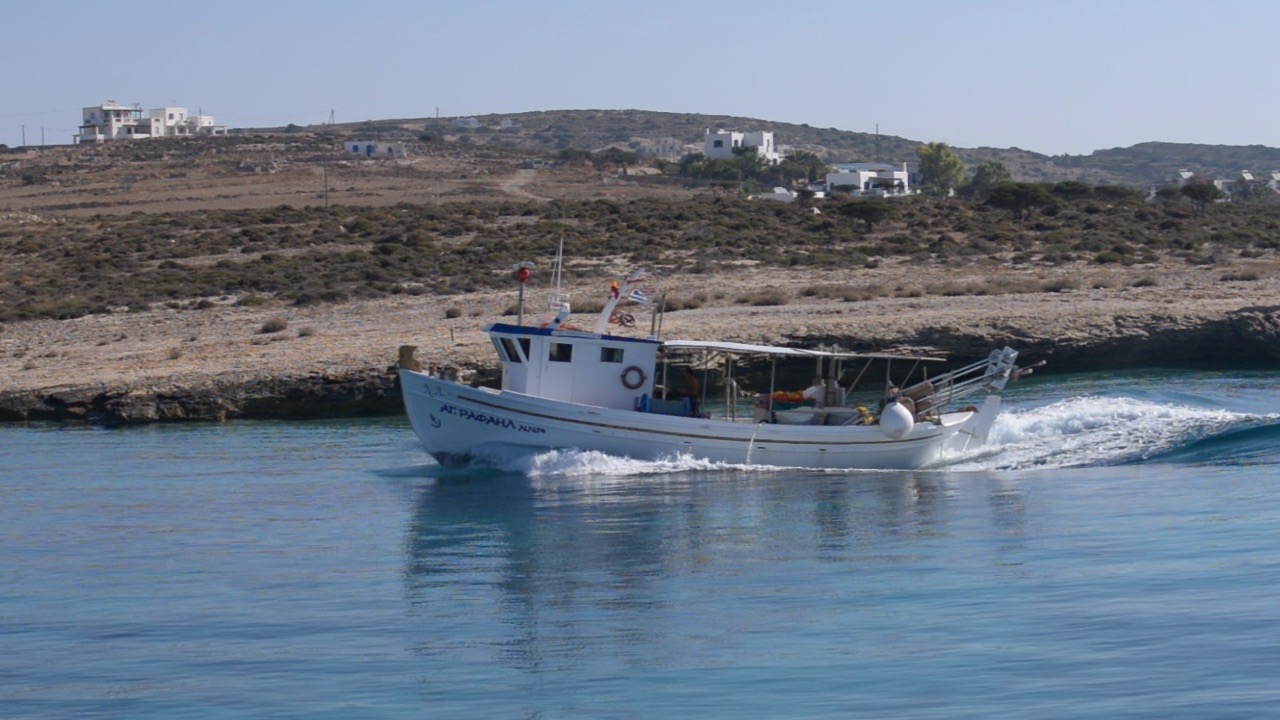}
    \end{subfigure}
    \begin{subfigure}{0.147\textwidth}
        \includegraphics[width=\textwidth]{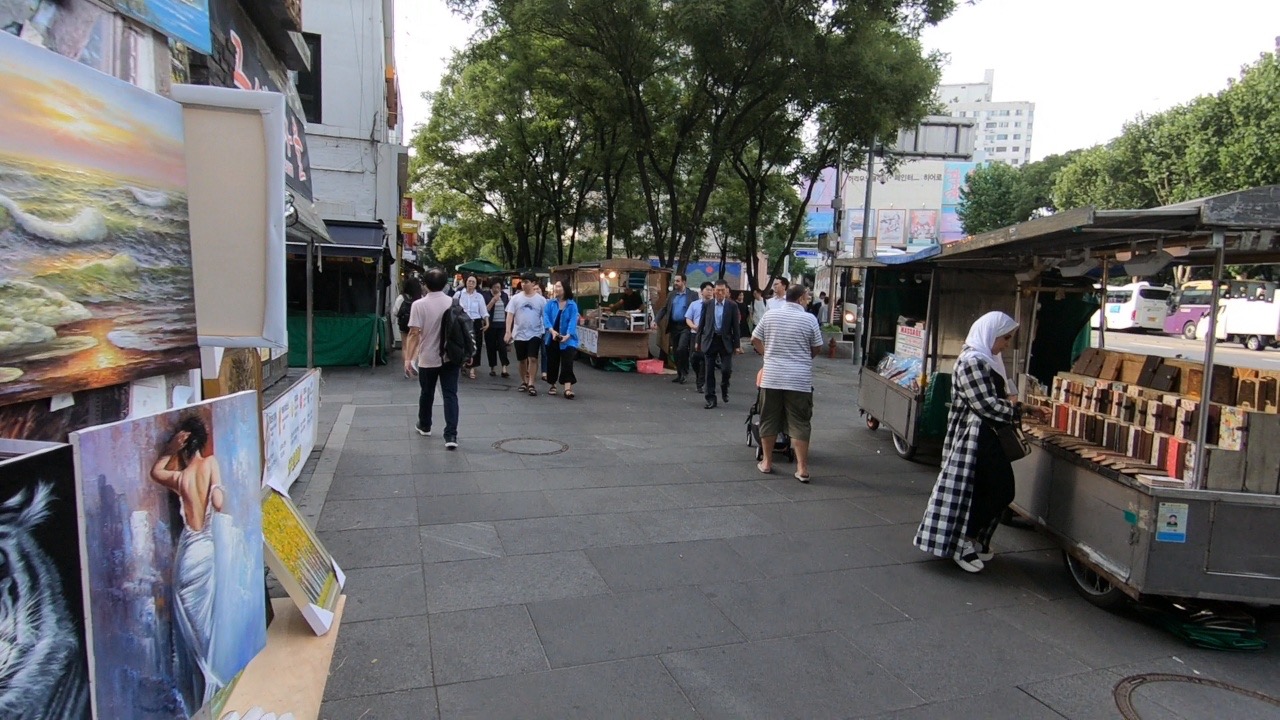}
    \end{subfigure}
    \\
    \begin{subfigure}{0.147\textwidth}
        \includegraphics[width=\textwidth]{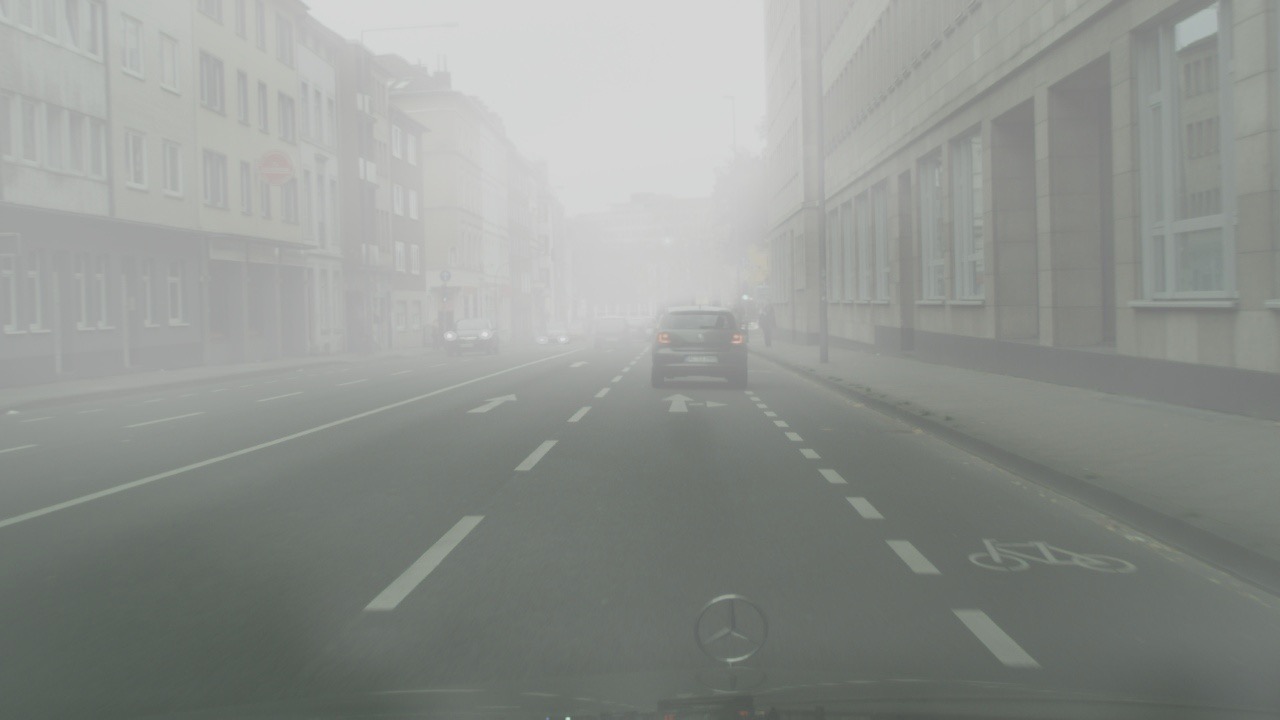}
    \end{subfigure}
    \begin{subfigure}{0.147\textwidth}
        \includegraphics[width=\textwidth]{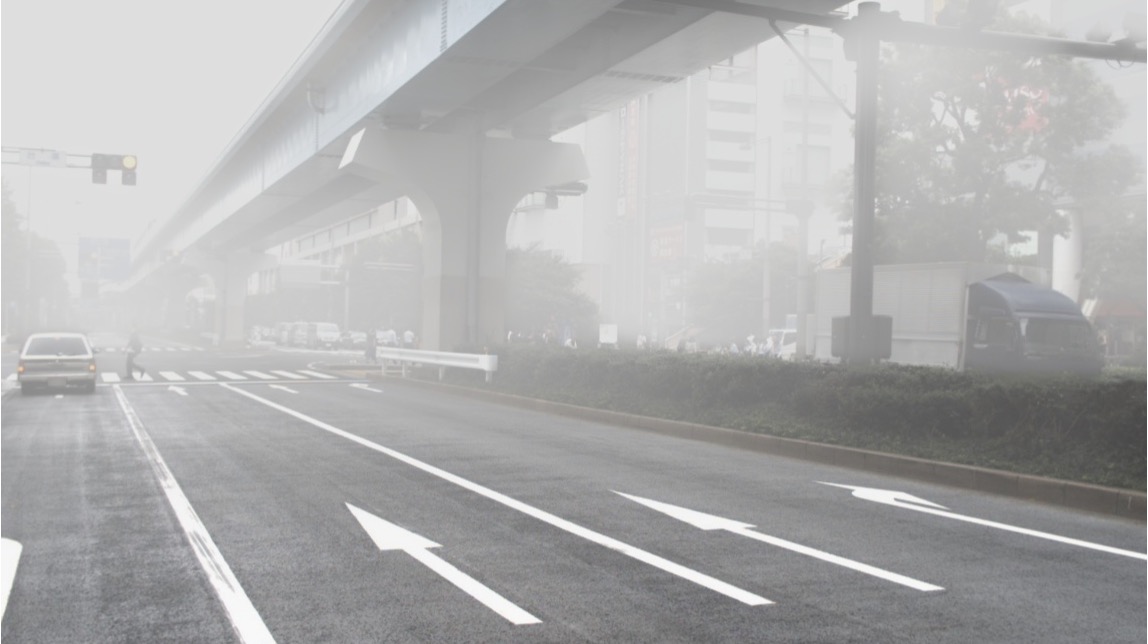}
    \end{subfigure}
    \begin{subfigure}{0.147\textwidth}
        \includegraphics[width=\textwidth]{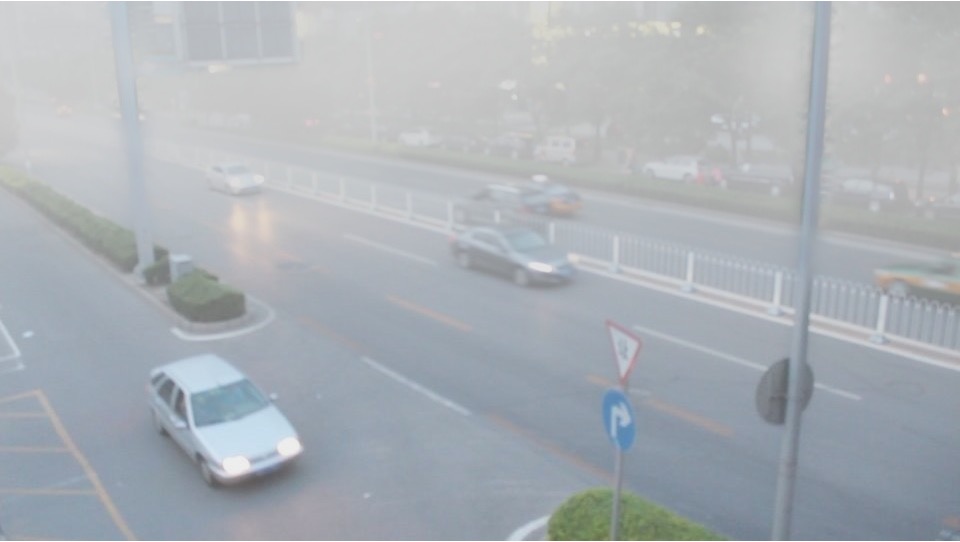}
    \end{subfigure}
    \begin{subfigure}{0.147\textwidth}
        \includegraphics[width=\textwidth]{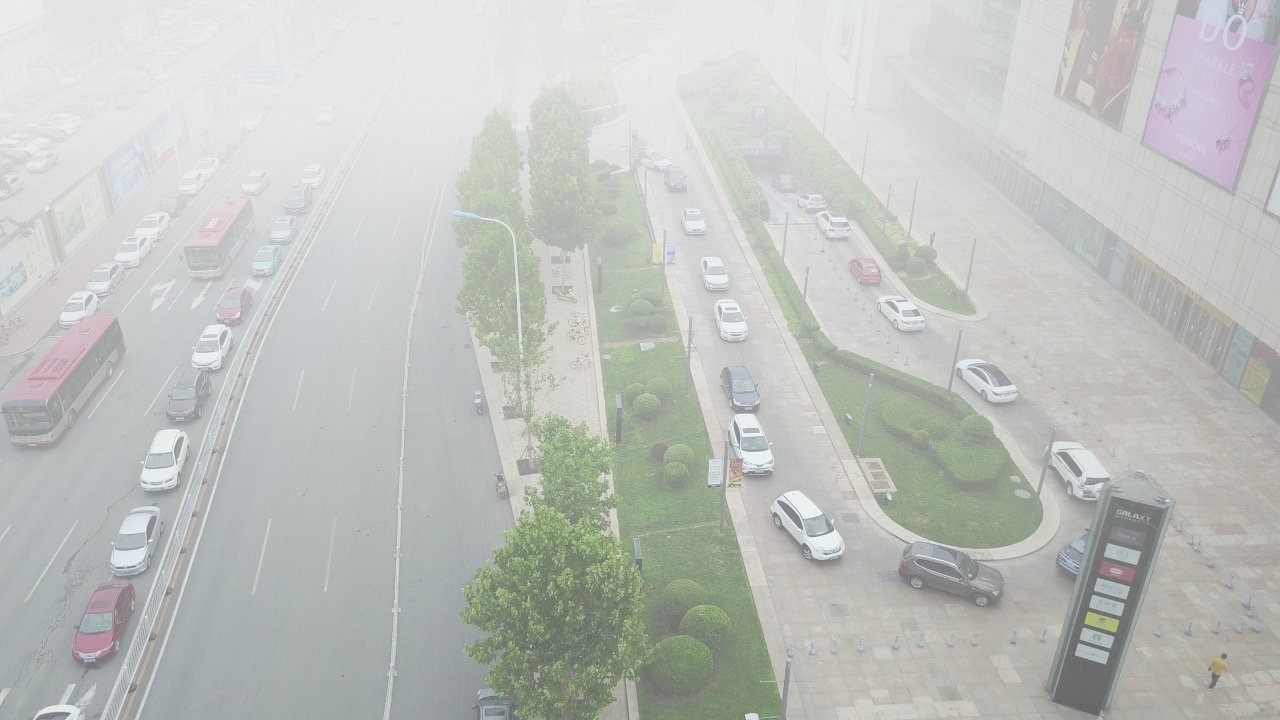}
    \end{subfigure}
    \begin{subfigure}{0.147\textwidth}
        \includegraphics[width=\textwidth]{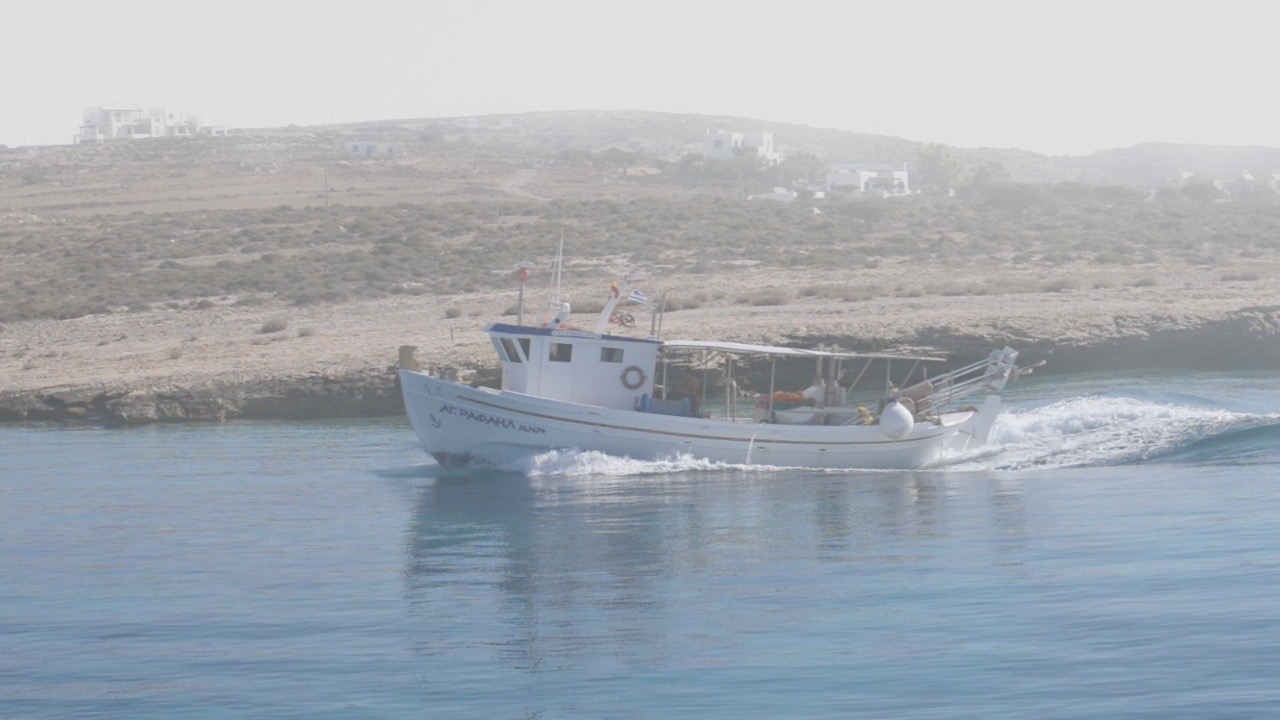}
    \end{subfigure}
    \begin{subfigure}{0.147\textwidth}
        \includegraphics[width=\textwidth]{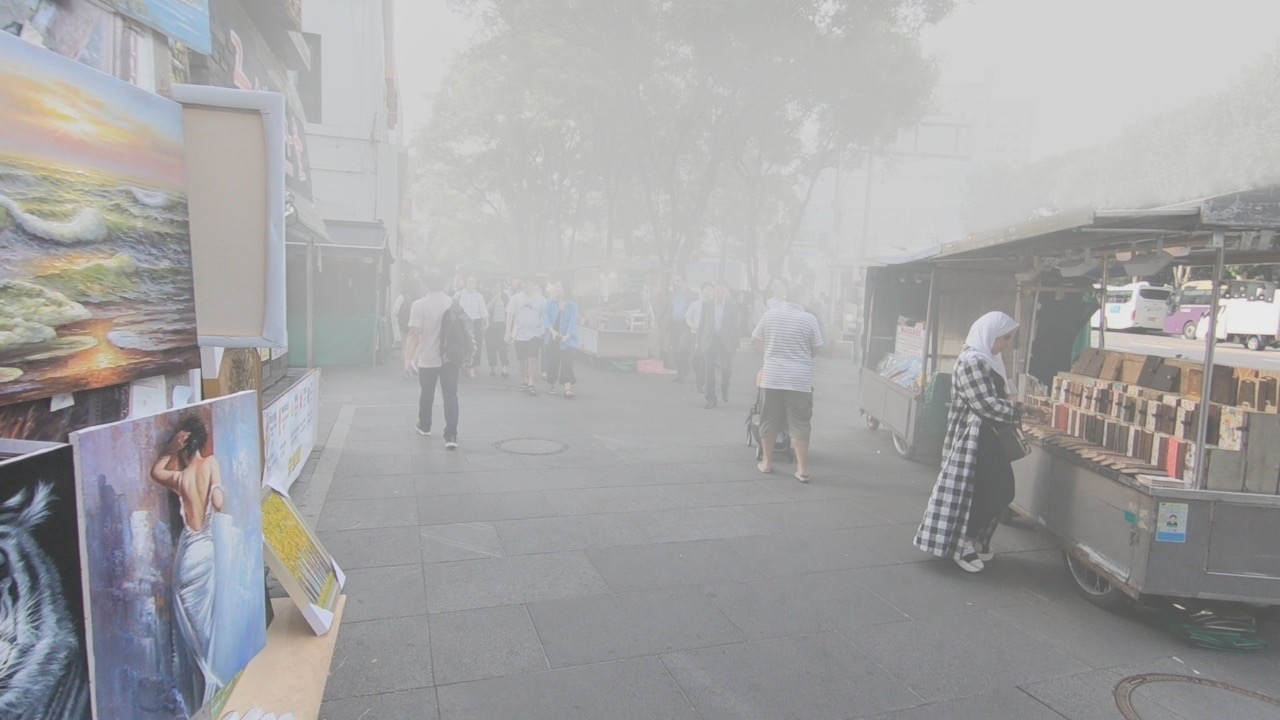}
    \end{subfigure}
    \\
    \begin{subfigure}{0.147\textwidth}
        \includegraphics[width=\textwidth]{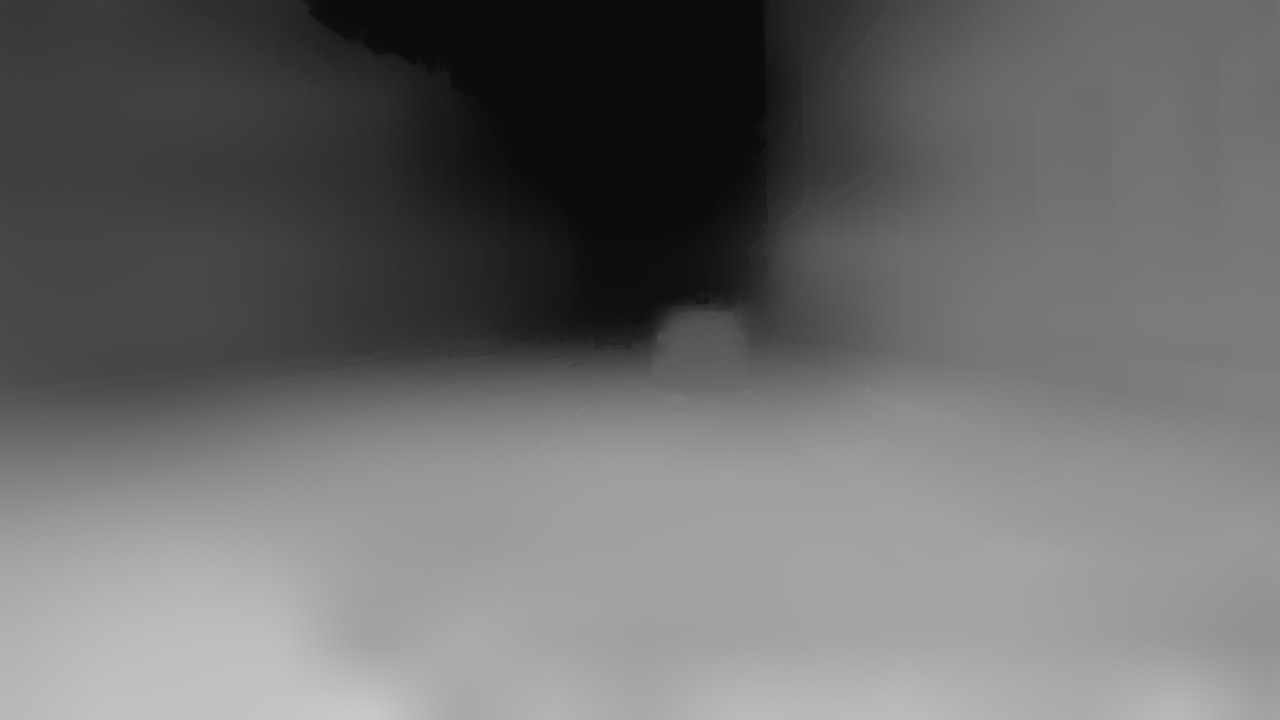}
        \caption{Cityscapes}
    \end{subfigure}
    \begin{subfigure}{0.147\textwidth}
        \includegraphics[width=\textwidth]{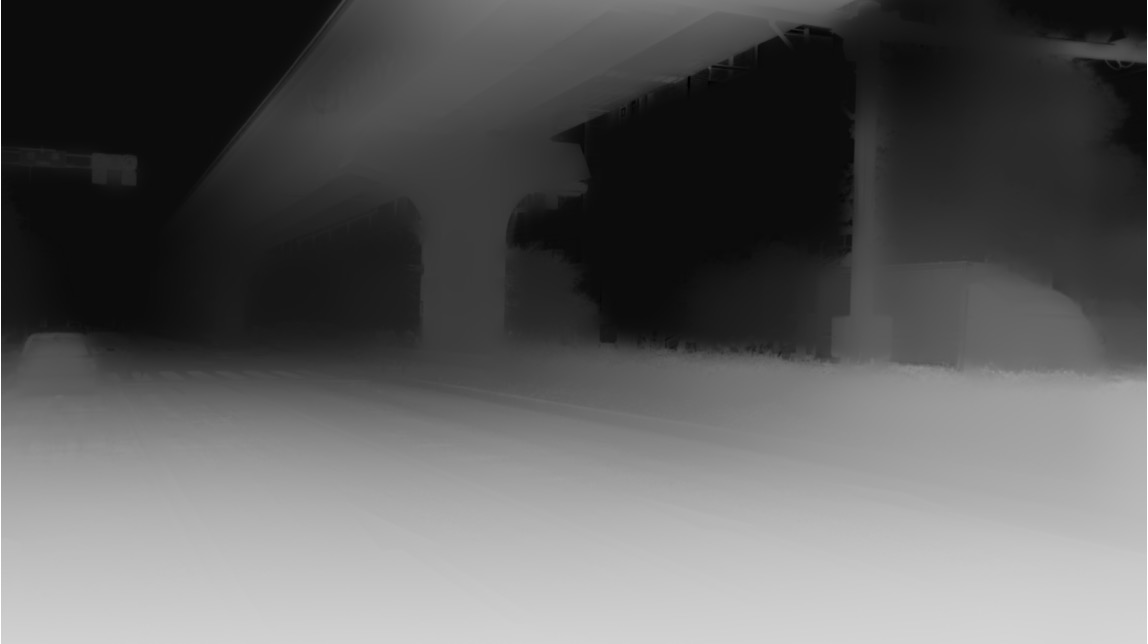}
        \caption{DDAD}
    \end{subfigure}
    \begin{subfigure}{0.147\textwidth}
        \includegraphics[width=\textwidth]{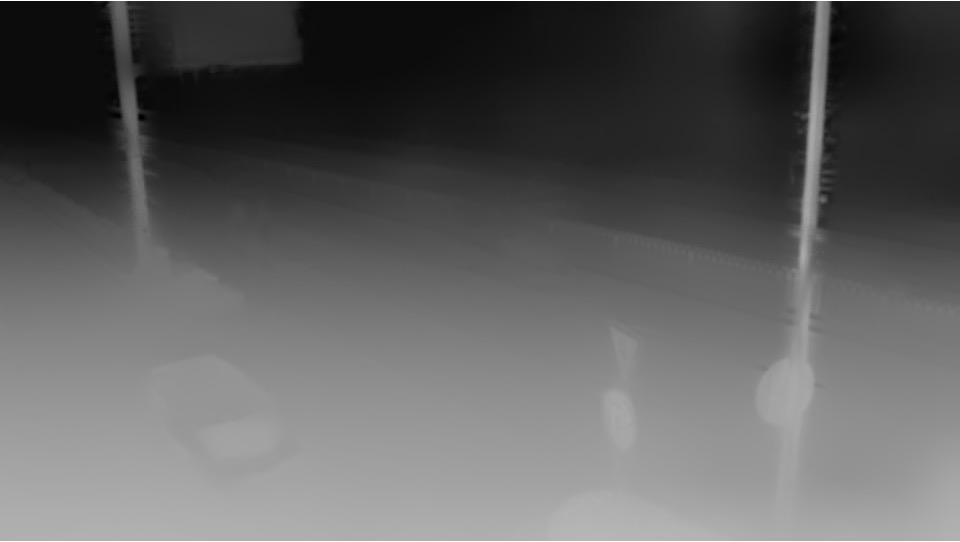}
        \caption{UA-DETRAC}
    \end{subfigure}
    \begin{subfigure}{0.147\textwidth}
        \includegraphics[width=\textwidth]{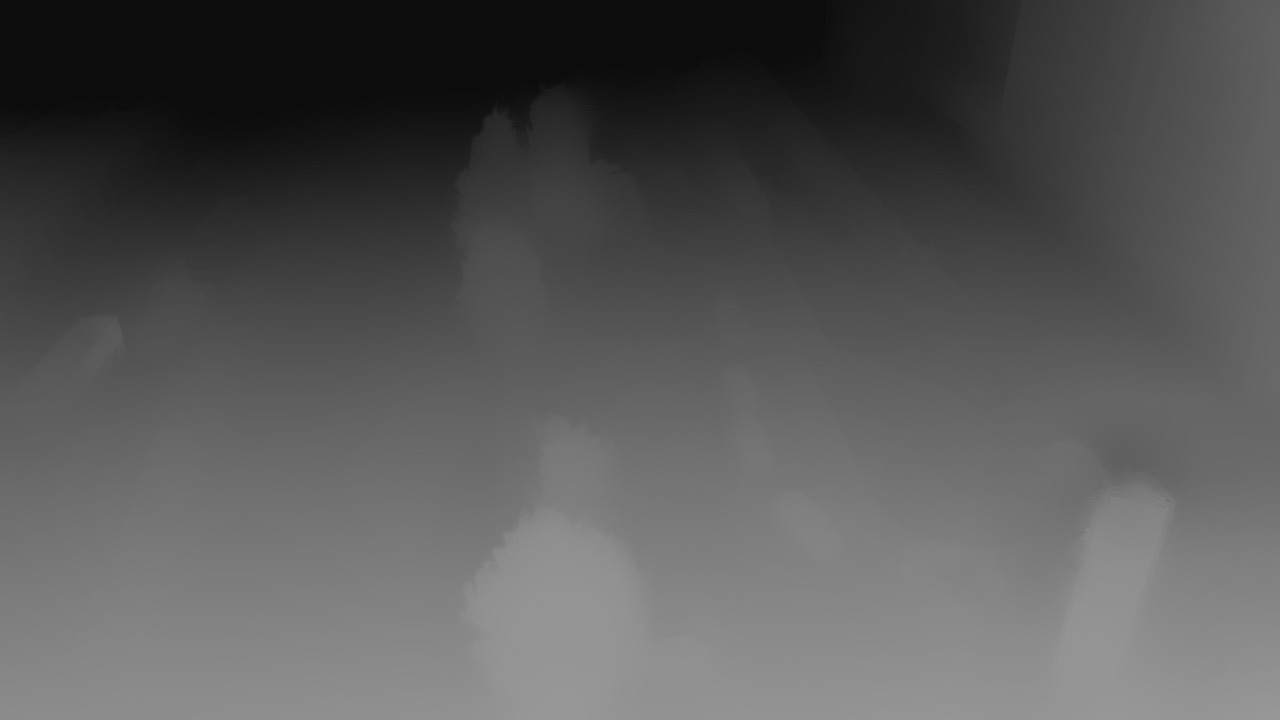}
        \caption{VisDrone}
    \end{subfigure}
    \begin{subfigure}{0.147\textwidth}
        \includegraphics[width=\textwidth]{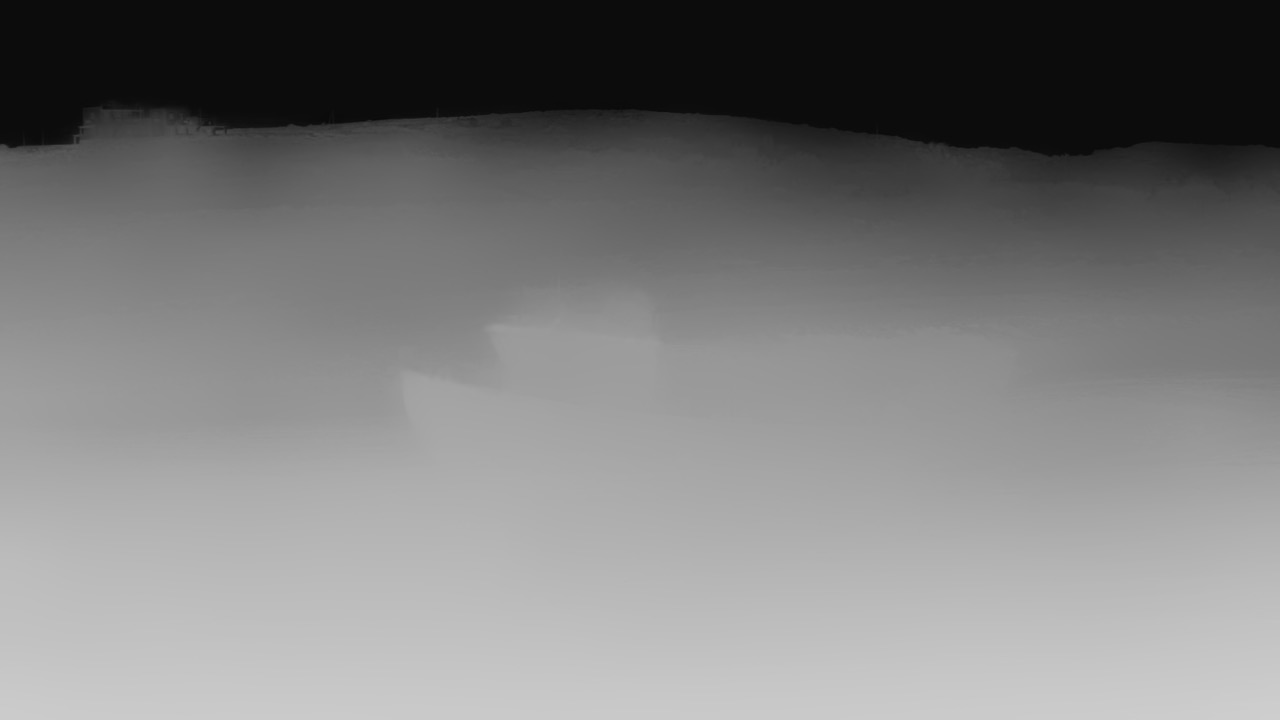}
        \caption{DAVIS}
    \end{subfigure}
    \begin{subfigure}{0.147\textwidth}
        \includegraphics[width=\textwidth]{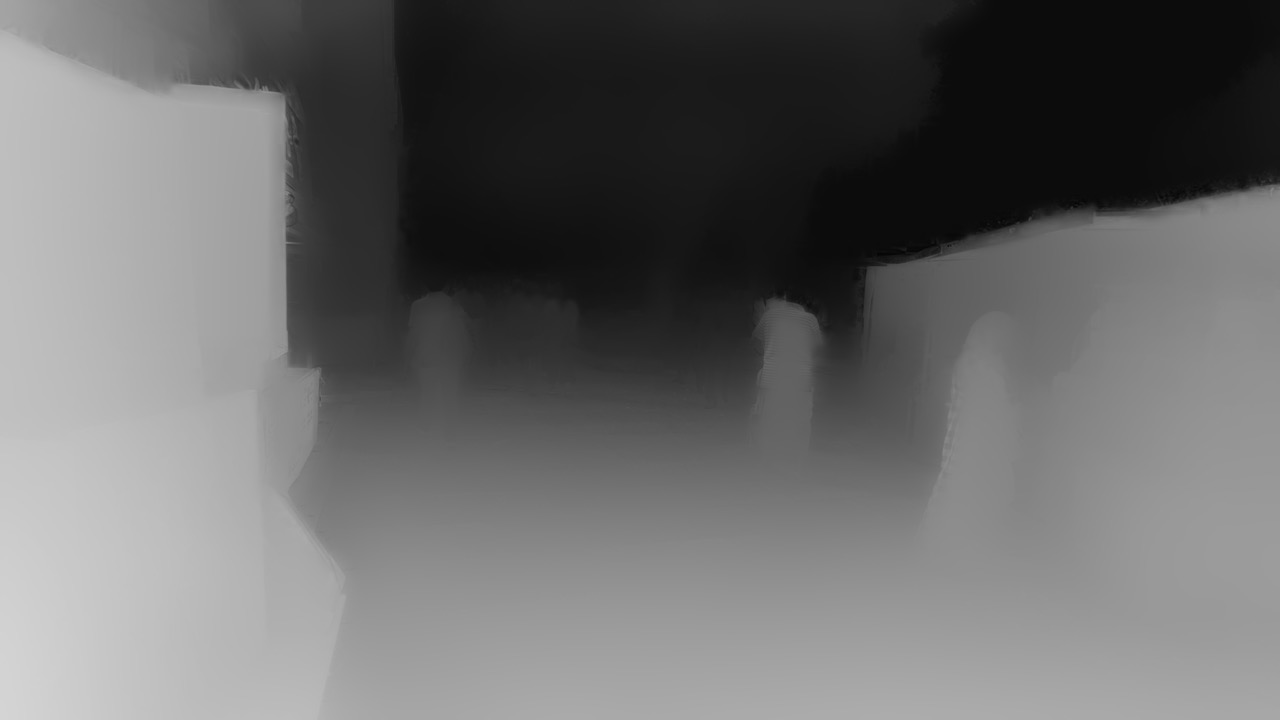}
        \caption{REDS}
    \end{subfigure}
    \\
    \vspace{-3mm}
    \caption{Example ground-truth images (the first row), synthetic hazy images (the second row), and transmission maps (the last row) in our HazeWorld dataset.}
    \label{fig:hazeworld}
    \vspace{-6mm}
\end{figure*}

\noindent \textbf{Image dehazing.}
Single-image dehazing has been widely studied in computer vision and computer graphics.
Early methods rely on the atmospheric scattering model and physical priors~\cite{he2010single,berman2016non}.
%
%
Later, deep learning-based methods show superior performance by leveraging large numbers of clear/hazy images~\cite{li2018benchmarking,ancuti2020nh}.
These methods either predict the components of the haze physical model~\cite{cai2016dehazenet,ren2016single,li2017aod,zhang2018densely} or directly restore the haze-free images in an image-to-image translation manner~\cite{li2018single,qu2019enhanced} using convolutional neural networks (CNNs).
%
Recent works propose more advanced network and module designs to improve the dehazing performance~\cite{deng2019deep,liu2019griddehazenet,chen2019gated,qin2020ffa,dong2020multi,liu2021synthetic,guo2022image,song2023vision}.
%
However, applying image dehazing methods to videos leads to discontinuous results since the temporal information is simply ignored.

\begin{figure}
    \centering
    \captionsetup[subfigure]{justification=centering}
    \includegraphics[width=0.98\hsize]{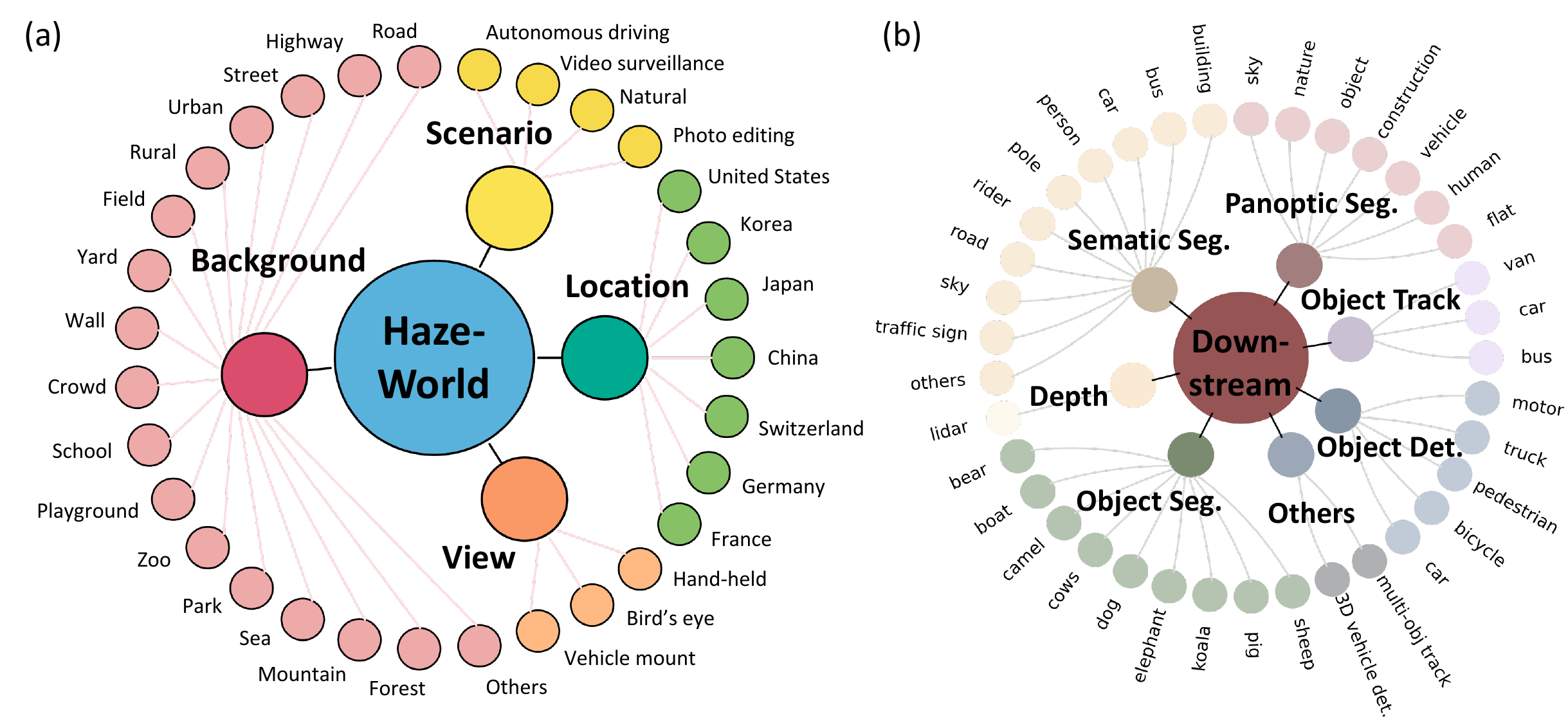}
    \vspace{-3mm}
    \caption{Dataset analysis of our HazeWorld, which contains diverse scenarios and supports various downstream evaluations.}
    \label{fig:analysis}
    \vspace{-5mm}
\end{figure}

\vspace{2mm}
\noindent
\textbf{Video dehazing.}
Video dehazing methods leverage temporal information from the adjacent frames to enhance the restoration quality.
Early methods mainly focus on post-processing to generate temporally consistent results by refining transmission maps and suppressing artifacts~\cite{kim2013optimized,chen2016robust} or joint estimating depths from videos~\cite{li2015simultaneous}.
Li~\etal~\cite{li2018end} present a CNN to optimize dehazing and detection in videos end-to-end.
Ren~\etal~\cite{ren2018deep} use semantic information to regularize the estimated transmission and to improve the video dehazing performance.
%
%
More recently, Zhang~\etal~\cite{zhang2021learning} collect a real indoor video dehazing dataset~(REVIDE) and present a confidence-guided and improved deformable network.
Liu~\etal~\cite{liu2022phase} design a phase-based memory network for video dehazing.
Additionally, a neural compression-based method~\cite{huang2022neural} for video restoration shows better performance on REVIDE.
However, these methods are mainly trained and evaluated in indoor scenes, and their performance in complex outdoor scenarios is limited.

\vspace{2mm}
\noindent
\textbf{Video alignment.}
Alignment aims at obtaining spatial transformation and pixel-wise correspondence from the misaligned frames.
Video restoration methods rely on explicit optical flow estimation~\cite{ranjan2017optical} to align the adjacent images/features~\cite{chan2021basicvsr,huang2022neural,yang2019frame}.
Other methods~\cite{tian2020tdan,wang2019edvr} leverage deformable convolutions~\cite{dai2017deformable} to learn the offsets for feature alignment.
These methods usually perform the frame-to-frame alignment.
More recently, attention~\cite{vaswani2017attention,dosovitskiy2020image,xia2022vision} with a large receptive field has been used together with the optical flow for feature alignment~\cite{cao2021video,lin2022flow,liang2022recurrent}.
Besides, STTN~\cite{kim2018spatio} also considers multiple frames but only transforms the input images at one space-time range.


\section{HazeWorld Dataset}

Since the current video dehazing datasets are mostly collected in indoor scenes, we construct a large-scale synthetic outdoor video dehazing dataset named HazeWorld, with example frames shown in~\cref{fig:hazeworld}.

\vspace{1.5mm}
\noindent
\textbf{Data collection.}
The original videos of HazeWorld are from six existing datasets, \ie, Cityscapes~\cite{cordts2016cityscapes}, DDAD~\cite{guizilini20203d}, UA-DETRAC~\cite{wen2020ua}, VisDrone~\cite{zhu2021detection}, DAVIS~\cite{pont20172017}, and REDS~\cite{nah2019ntire}, resulting in 1,271 haze-free videos.
%
We use the atmospheric scattering model~\cref{eq:physical model} to synthesize hazy videos.
The robust video depth estimation method~\cite{kopf2021robust} is used to obtain temporally consistent depth maps.
%
%
We follow~\cite{sakaridis2018semantic,bijelic2020seeing} and choose $\beta \in \{0.005,0.01,0.02,0.03\}$ to generate transmission $t$, and 
randomly select $A\in [0.75,1.0]$ for each video.
%
We split 1,271 haze-free videos into training (897 videos) and testing (374 videos) sets.
Overall, we obtain 3,588 and 1,496 hazy synthetic videos with four $\beta$ of around 240,000 and 86,000 frames in training and testing sets, respectively.

\vspace{1.5mm}
\noindent
\textbf{Dataset analysis.}
As shown in \cref{fig:analysis}, our dataset contains diverse real-world scenarios, which enables us to assess dehazing performance on various outdoor applications, such as autonomous driving~\cite{cordts2016cityscapes,guizilini20203d}, video surveillance~\cite{wen2020ua,zhu2021detection}, and photo editing~\cite{nah2019ntire}.
Further, the original datasets contain the labels of multiple video and image downstream scene understanding tasks, \eg, video panoptic segmentation~\cite{kim2020video}, object segmentation~\cite{pont20172017}, depth estimation~\cite{guizilini20203d}, and image semantic segmentation~\cite{cordts2016cityscapes}. Thus we can evaluate the effectiveness of dehazing on high-level vision tasks.



    







\section{Methodology}

\begin{figure*}[t]
    \centering
    \includegraphics[width=0.9\hsize]{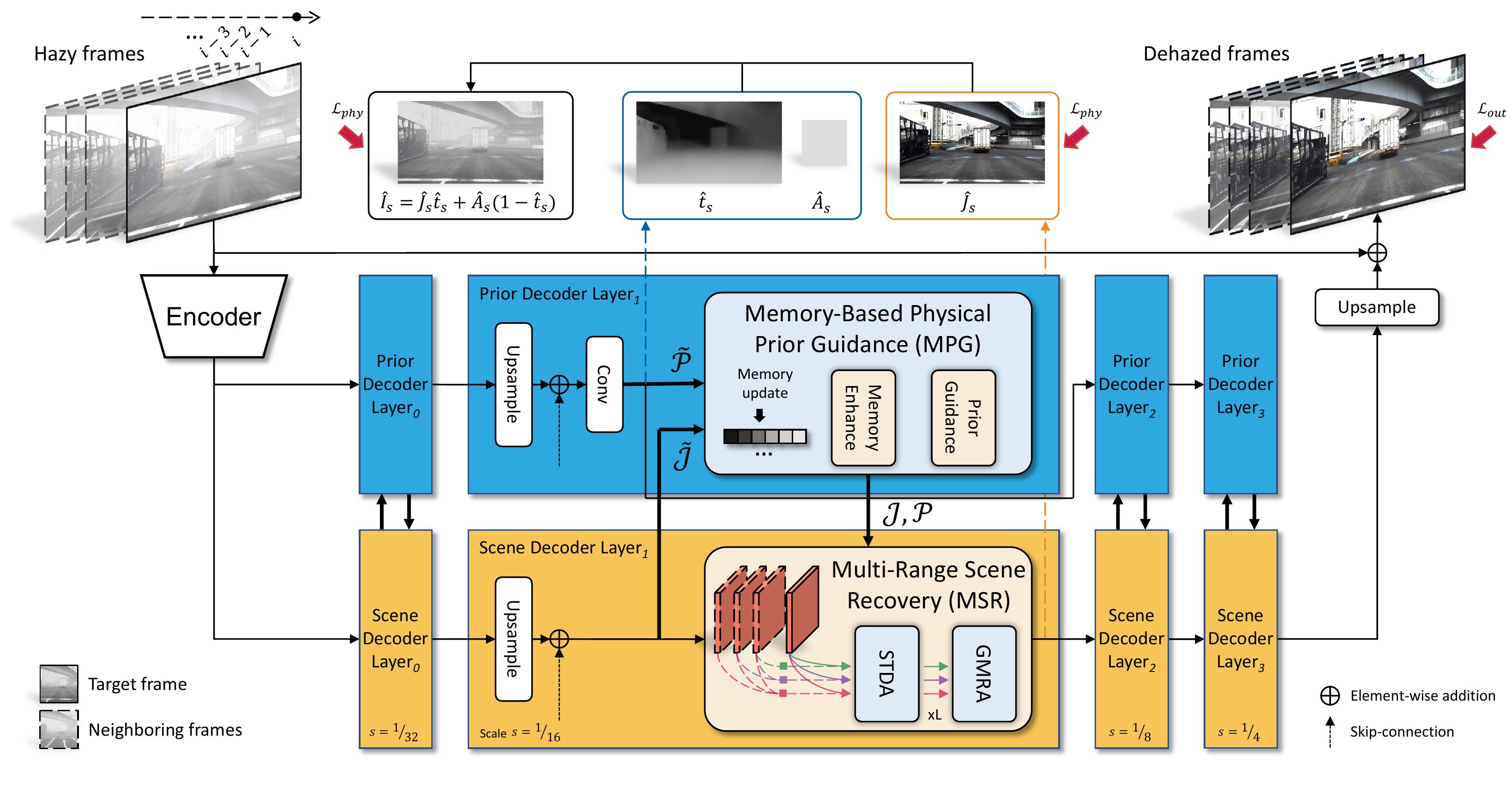}
    \vspace{-2.5mm}
    \caption{The overall framework of our MAP-Net for video dehazing. MAP-Net is a U-Net-like structure that mainly contains an encoder, a prior decoder, and a scene decoder. Features are processed interactively in the prior decoder and scene decoder, which jointly perform feature disentanglement. The former produces the prior guidance with a memory, and the latter recovers the scene recurrently.}
    \vspace{-3mm}
    \label{fig:overview}
\end{figure*}

\subsection{Overall Framework}

\cref{fig:overview} illustrates the overall framework of the proposed MAP-Net, which
%
%
is a U-Net-like structure that mainly consists of an encoder, a prior decoder, and a scene decoder.
%
A common image backbone, \eg, ConvNeXt~\cite{liu2022convnet}, is used as the feature encoder, which extracts the multi-scale feature maps.
%
At each scale, features are processed interactively in the prior decoder layer and scene decoder layers.
The initial prior feature $\tilde{\mathcal{P}}$ and initial scene feature $\tilde{\mathcal{J}}$ are first fed into a Memory-based Physical prior Guidance (MPG) module (see~\cref{sec:MPG}), which aims to obtain the memory-enhanced prior feature $\mathcal{P}$ and the prior-guided scene feature $\mathcal{J}$.
Then, $\mathcal{P}$ and $\mathcal{J}$ are fed into a Multi-range Scene radiance Recovery (MSR) module (see~\cref{sec:MSR})
, which is to obtain the feature for the haze-free scene by aligning and aggregating recurrent temporal features from the adjacent frames.
%
%
%
The prior decoder and scene decoder jointly perform feature disentanglement according to the physical model.

Specifically, the prior decoder learns the prior-related feature by predicting the transmission and atmospheric light, and the scene decoder generates the scene radiance.
The intermediate components are obtained using separate prediction heads and reconstructing the hazy input via~\cref{eq:physical model}, which are supervised by a physical model disentanglement loss.
%
Moreover, pixel shuffle layers~\cite{shi2016real} are used to upsample the features in two decoders and the output from the last scene decoder layer.
%
Lastly, residual prediction is used to produce the final dehazed result.

%

\subsection{Memory-Based Physical Prior Guidance}
\label{sec:MPG}

\begin{figure}
    \centering
    \includegraphics[width=\hsize]{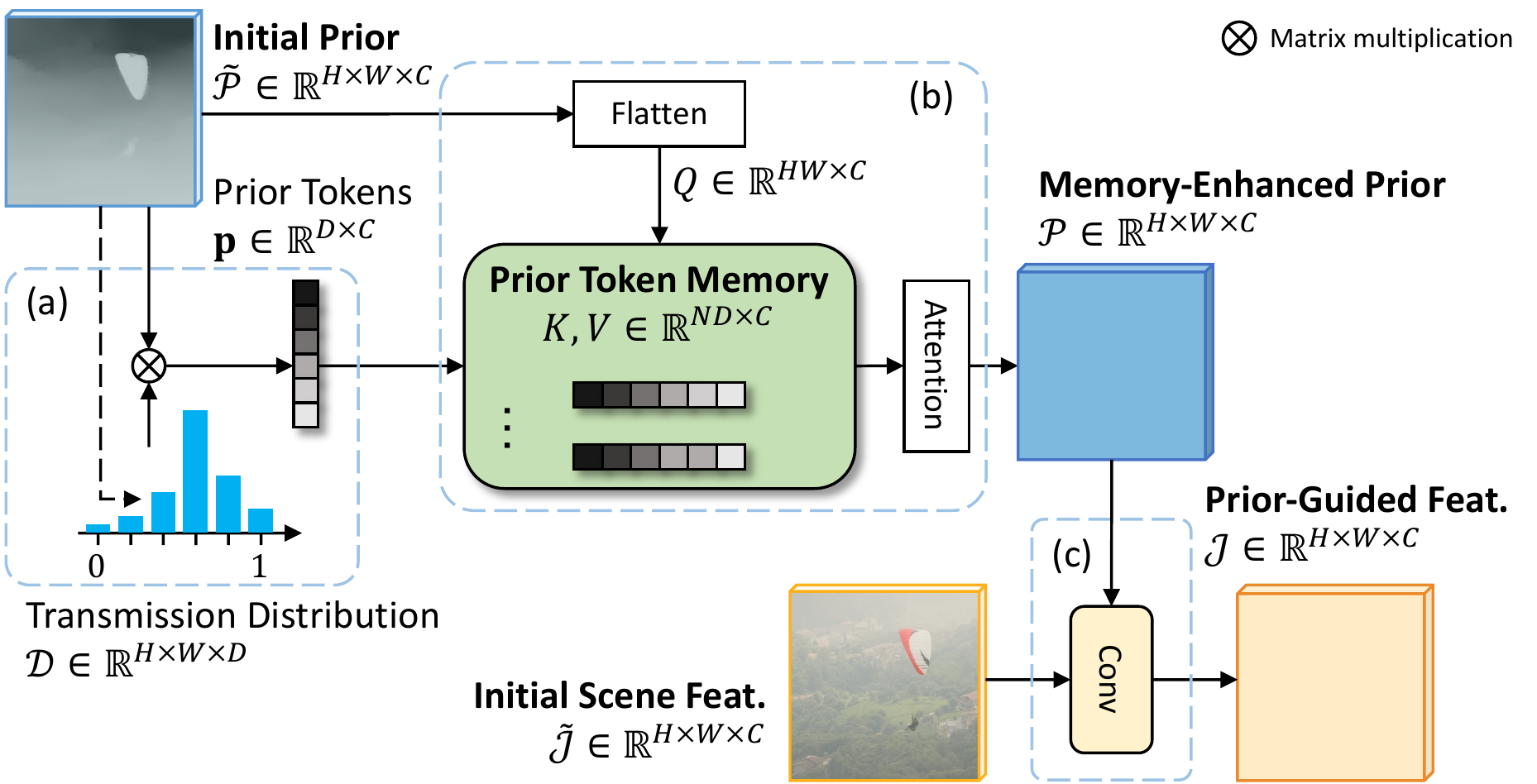}
    \caption{The illustration of the Memory-based physical Prior Guidance (MPG) module, which has (a) physical prior compression, (b) memory-enhanced prior, and (c) prior feature guidance.}
    \label{fig:mpg}
    \vspace{-3mm}
\end{figure}

We design a Memory-based Physical prior Guidance (MPG) module to enhance the scene recovery by encoding haze prior-related features into the long-range memory.
%
%
\cref{fig:mpg} shows the architecture of MPG with three parts.

%
%
%
%
%

\vspace{2mm}
\noindent
\textbf{Physical prior compression.}
%
%
The initial prior feature $\tilde{\mathcal{P}} \in \mathbb{R}^{H \times W \times C}$ is implicitly learned using several convolution layers on the upsampled features to predict the transmission map and atmospheric light.
$H$, $W$, and $C$ denote the feature height, width, and channel size.
%
Since we aim to save the physical priors at different times in the memory, we need to compress the size of each prior to reduce memory space.
To achieve this, we first perform a discretization operation on the prior feature by using categorical classification~\cite{fu2018deep,huang2022monodtr} and then normalize the results via the Softmax function.
%
%
Specifically, from the initial prior, we generate the transmission distribution map $\mathcal{D} \in \mathbb{R}^{H \times W \times D}$, where $D$ is the number of transmission categories.
%
%
%
After that, we perform matrix multiplication between the initial prior $\tilde{\mathcal{P}}$ and the transmission distribution map $\mathcal{D}$ to obtain the value $\textbf{p} \in \mathbb{R}^{D \times C}$, which is the compressed prior token. 

\begin{figure*}[ht]
    \centering
    \includegraphics[width=\hsize]{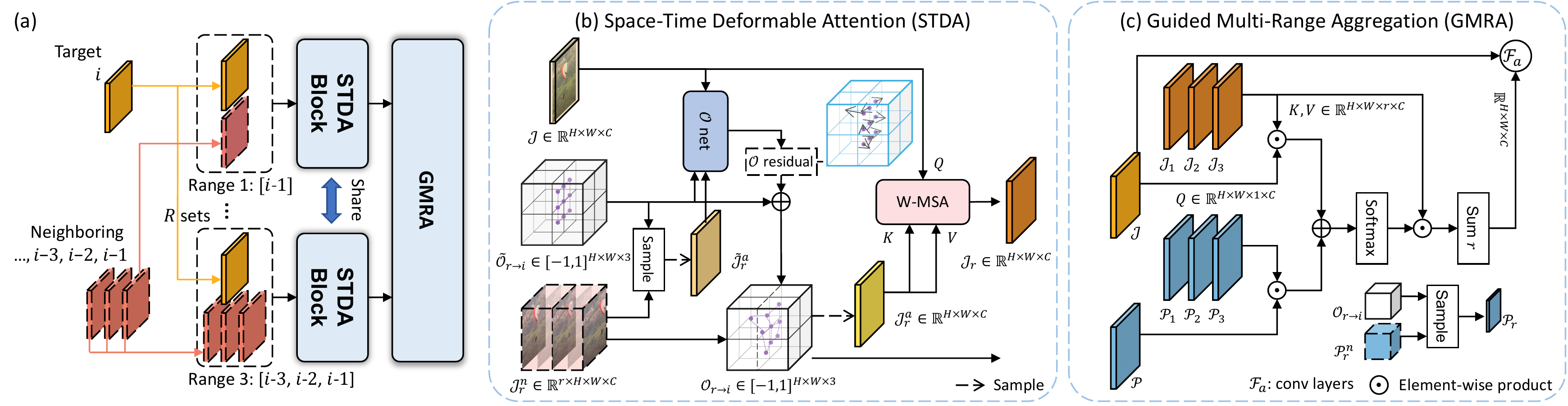}
    \caption{The illustration of our Multi-range Scene radiance Recovery (MSR) module. (a) MSR aligns the features of neighboring frames into multiple sets with different ranges. (b) The space-time deformable attention (STDA) block aligns the features of different ranges to the target frame. (c) The guided multi-range aggregation (GMRA) block aggregates the aligned features from multiple sets.}
    \label{fig:msr}
    \vspace{-3mm}
\end{figure*}

\vspace{2mm}
\noindent
\textbf{Memory-enhanced prior.}
After obtaining the compressed prior token $\textbf{p}$, we formulate a prior token memory by saving multiple prior tokens at different time slots.
Then, we obtain the feature vectors $K$ and $V$ with the dimension of $ \mathbb{R}^{ND \times C}$, where $N$ denotes the number of prior tokens.
Hence, we are able to record the historical haze information in video sequences.
To perform the interaction between the current haze information and the history information encoded in prior token memory, we adopt the attention operation to read the memory information:
\begin{equation}
	\textrm{Attention}(Q, K, V) = \textrm{softmax}(\frac{QK^T}{\sqrt{c}})V \ ,
\end{equation}
where the query $Q$ is obtained by flatting the initial prior $\tilde{\mathcal{P}}$, and $c$ is the normalization factor, which is the dimension of $Q$ and $K$.
By doing so, we obtain the final memory-enhanced prior $\mathcal{P} = \textrm{Attention}(Q, K, V)$.

\vspace{2mm}
\noindent
\textbf{Prior feature guidance.}
The prior-guided scene feature $\mathcal{J}$ is obtained using several convolution layers, which take the concatenation of the memory-enhanced prior feature $\mathcal{P}$ and the initial scene feature $\tilde{\mathcal{J}}$ as input.
Hence, the prior is integrated for scene recovery.

\subsection{Multi-Range Scene Radiance Recovery}
\label{sec:MSR}
The Multi-range Scene radiance Recovery module (MSR) aims to capture space-time dependencies in multiple space-time ranges.
~\cref{fig:msr} shows the detailed structure of our MSR, which aligns the features of adjacent frames $\mathcal{J}_{[{i-1}, {i-2}, {i-3}, ...]}$ into multiple sets with different ranges, \ie, $\mathcal{J}_{\{[{i-1}], [{i-1},{i-2}], [{i-1},{i-2},{i-3}], ...\}}$, to explore the temporal haze clues in various time intervals.
As shown in~\cref{fig:msr}~(a), the concatenated features with different ranges are sent to the shared space-time deformable attention block (STDA), which warps the features to the target frame.
%
%
After that, we formulate a guided multi-range aggregation block (GMRA) to aggregate the aligned features from multiple sets with the guidance of prior features.


%
%
%

\vspace{-1mm}
\subsubsection{Space-Time Deformable Attention}
\label{sec:stda}
As shown in~\cref{fig:msr}~(b), the space-time deformable attention (STDA) block aligns the concatenated features of the adjacent frames $\mathcal{J}_r^n=\mathcal{J}_{[{i-1},...,{i-r}]} \in \mathbb{R}^{r \times H \times W \times C}$ for each range $r\in\{1,...,R\}$ towards the target frame feature $\mathcal{J}$.
The output of the STDA block is the range feature $\mathcal{J}_r \in \mathbb{R}^{H \times W \times C}$.
%
Here, we further learn a space-time flow, 
which is used to capture the correspondence from the previous frames to the current frame.
The input space-time flow of the current STDA block is $\tilde{\mathcal{O}}_{r \rightarrow i}$, which is gradually refined in this block to produce the output space-time flow $\mathcal{O}_{r \rightarrow i}$, following SPy-Net~\cite{ranjan2017optical}.
%


%
%

Specifically, given the concatenated features $\mathcal{J}_r^n$ and a normalized initial space-time flow $\tilde{\mathcal{O}}_{r \rightarrow i} \in [-1,1]^{H \times W \times 3}$, 
we first compute the initial aligned feature map $\tilde{\mathcal{J}}_r^a \in \mathbb{R}^{H \times W \times C}$ as follows:
\begin{equation}
    \label{eq:sample}
    \tilde{\mathcal{J}}_r^a = \mathcal{S}(\mathcal{J}_r^n, \tilde{\mathcal{O}}_{r \rightarrow i}) \ ,
\end{equation}
%
where $\mathcal{S}$ denotes the differentiable space-time sampling operation~\cite{kim2018spatio}.
Note that the third dimension of $\tilde{\mathcal{O}}_{r \rightarrow i} \in [-1,1]^{H \times W \times 3}$ is three, which means the space-time flow capture locations on both spatial domain and time slot.
Then, we obtain the refined space-time flow $\mathcal{O}_{r \rightarrow i}$ by computing the flow offset residual:
%
\begin{equation}
    \mathcal{O}_{r \rightarrow i} = \mathcal{F}_o([\mathcal{J}, \tilde{\mathcal{J}}_r^a, \tilde{\mathcal{O}}_{r \rightarrow i}]) + \tilde{\mathcal{O}}_{r \rightarrow i} \ ,
\end{equation}
where $\mathcal{F}_o$ is a lightweight offset network composed of convolution layers.
%
Finally, we obtain the aligned feature $\mathcal{J}_r^a$ using \cref{eq:sample} with $\mathcal{O}_{r \rightarrow i}$ as input instead.

Cross-attention~\cite{vaswani2017attention} is used to extract the temporal information from the aligned feature.
The feature $\mathcal{J}$ is used as the query $Q = \mathcal{J} U_q$ for the target frame, and the aligned feature $\mathcal{J}_r^a$ is used as the key and value $[K, V] = \mathcal{J}_r^a U_{kv}$, where $U_q \in \mathbb{R}^{C \times C}, U_{kv} \in \mathbb{R}^{C \times 2C}$ are learnable projection matrices.
Finally, the range feature $\mathcal{J}_r$ is computed as:
\begin{equation}
    \mathcal{J}_r = \textrm{W-MSA} (Q, K, V) \ ,
\end{equation}
where the window multi-head self attention (W-MSA)~\cite{liu2021swin} is leveraged for efficient computation; see~\cref{fig:msr}~(b).
Note that we further adopt the feed-forward network (FFN) to process $\mathcal{J}_r$  after the W-MSA, following~\cite{liu2021swin}. 
%

\begin{table*}[t]
\caption{\textbf{Quantitative comparison with state-of-the-art methods on our HazeWorld dataset.} \textbf{Bold} and \underline{underline} indicate the best and the second-best performance, respectively.}
\label{tab:exp-hazeworld}
\centering
\begin{adjustbox}{width=1\textwidth}
\footnotesize
\begin{tabular}{lccccccccc}
\hline
Method          & DCP~\cite{he2010single} & AOD~\cite{li2017aod} & GDN~\cite{liu2019griddehazenet} & DM2F~\cite{deng2019deep} & FFA~\cite{qin2020ffa} & MSBDN~\cite{dong2020multi} & UHD~\cite{zheng2021ultra} & AECR~\cite{wu2021contrastive} & Dehamer~\cite{guo2022image} \\
\hline
PSNR $\uparrow$ & 16.49                   & 15.46                & 22.80                           & 24.54                    & 22.11                 & 23.70                      & 19.43                     & 22.04                         & 22.92                       \\
SSIM $\uparrow$ & 0.8126                  & 0.7997               & 0.9217                          & 0.9130                   & 0.9007                & 0.8858                     & 0.7807                    & 0.9067                        & 0.9044                      \\
\hline
Method          & DehazeFormer~\cite{song2023vision} & EVD~\cite{li2018end} & VDH~\cite{ren2018deep} & CG-IDN~\cite{zhang2021learning} & FastDVD~\cite{tassano2020fastdvdnet} & EDVR~\cite{wang2019edvr} & NCFL~\cite{huang2022neural} & BasicVSR++~\cite{chan2021basicvsr++} & Our method      \\
\hline
PSNR $\uparrow$ & 25.44                              & 15.91                & 17.97                  & 25.25                           & 21.25                                & 22.91                    & 24.33                       & \underline{26.06}                    & \textbf{27.12}  \\
SSIM $\uparrow$ & \underline{0.9286}                 & 0.7968               & 0.7780                 & 0.9155                          & 0.8678                               & 0.9036                   & 0.9253                      & 0.9207                               & \textbf{0.9349} \\
\hline
\end{tabular}
\end{adjustbox}
\vspace{-2mm}
\end{table*}
\begin{table*}[t]
\caption{\textbf{Quantitative comparison on the REVIDE dataset~\cite{zhang2021learning}.} Other baseline results are provided by NCFL~\cite{huang2022neural} paper.}
\label{tab:exp-REVIDE}
\centering
\begin{adjustbox}{width=1\textwidth}
\small
\begin{tabular}{lcccccccccc}
\hline
Method          & DCP~\cite{he2010single} & GDN~\cite{liu2019griddehazenet} & MSBDN~\cite{dong2020multi} & FFA~\cite{qin2020ffa} & VDH~\cite{ren2018deep} & EDVR~\cite{wang2019edvr} & CG-IDN~\cite{zhang2021learning} & NCFL~\cite{huang2022neural} & BasicVSR++~\cite{chan2021basicvsr++} & Our method             \\
\hline
PSNR $\uparrow$ & 11.03                   & 19.69                           & 22.01                      & 16.65                 & 16.64                  & 21.22                    & 23.21                           & \underline{23.63}           & 21.68                                 & \textbf{24.16}   \\
SSIM $\uparrow$ & 0.7285                  & 0.8545                          & 0.8759                     & 0.8133                & 0.8133                 & 0.8707                   & 0.8836                          & \underline{0.8925}          & 0.8726                                & \textbf{0.9043}  \\
\hline
\end{tabular}
\end{adjustbox}
\vspace{-2mm}
\end{table*}

\vspace{-1mm}
\subsubsection{Guided Multi-Range Aggregation}
The aligned features of different ranges contain their specific space-time haze clues, and GMRA aggregates multi-range features under the guidance of prior features.
%
%
\cref{fig:msr}~(c) shows the detailed structure, where we compute the aggregation weights from two perspectives, \ie, scene radiance and physical prior.
First, the concatenated range features $\{\mathcal{J}_r\}_{r=1}^R$ are considered as the key and value, which are multiplied by the target frame query feature~$\mathcal{J}$. 
For each location, attention weights are computed along the range~($r$) dimension.
%
Then, the prior guidance is leveraged by computing its derived attention weights in the same way.
In specific, we consider the prior feature $\mathcal{P}$ as query and obtain the aligned prior features $\{\mathcal{P}_r\}_{r=1}^R$ as the attention key using Eq.~(\ref{eq:sample}) by taking the prior features $\mathcal{P}_r^n=\mathcal{P}_{[{i-1},...,{i-r}]}$ and the refined space-time flow $\mathcal{O}_{r \rightarrow i}$ as the inputs.
%
The final attention weight is the summation of the weights generated from the scene and prior aspects, followed by a Softmax normalization function for normalization.
Finally, multi-range values are aggregated by performing the final attention weight on the range features, which is further summed along the $r$ (range) dimension; see~\cref{fig:msr}~(c). 
%

\subsection{Loss Functions}
\label{sec:loss}
The overall loss $\mathcal{L}$ is the summation of an output loss $\mathcal{L}_{out}$, a physical model disentanglement loss $\mathcal{L}_{phy}$, and a flow loss $\mathcal{L}_{flow}$:
\begin{equation}
\label{eq:loss}
    \mathcal{L} = \mathcal{L}_{out}
    + {\lambda}_{phy} \mathcal{L}_{phy} + {\lambda}_{flow} \mathcal{L}_{flow},
\end{equation}
where ${\lambda}_{rec}, {\lambda}_{flow}$ are the weighting hyper-parameters.

The output loss $\mathcal{L}_{out}=\mathcal{L}_1(\hat{J}, J)$ supervises the final dehazed results $\hat{J}$ with the ground truth $J$.  
The physical model disentanglement loss $\mathcal{L}_{phy}=\sum_{s=0}^3 {2^{s-3} \mathcal{L}_1(\hat{I}_s, I_s)} + \mathcal{L}_1(\hat{J}_s, J_s)$ is to make the prior decoder and scene decoder learn the physical model-based components at each scale $s$ in the U-Net, by predicting $\hat{t}_s, \hat{A}_s$, and $\hat{J}_s$, and reconstructing input $\hat{I}_s$ using Eq.~\eqref{eq:physical model}.
Moreover, to make the STDA attend to informative regions, we use an unsupervised flow loss to regularize the learned space-time flow in~\cref{sec:stda}.
Specifically, the flow loss $\mathcal{L}_{flow}=\sum_{s=0}^3 \sum_{r=1}^R {2^{s-3} \mathcal{L}_1(\hat{J}_{sr}^a, J_s)}$ computes the difference between the warped image $\hat{J}_{sr}^a$ and the reference ground truth frame $J_s$ with the scale $s$ for each range $r$.
The warped image is obtained by~\cref{eq:sample} with the adjacent ground truth frames and the learned space-time flow as the inputs.
%

\section{Experimental Results}

\begin{figure*}
    \centering
    \captionsetup[subfigure]{labelformat=empty,justification=centering}
    %
    %
    \begin{subfigure}{0.12\textwidth}
        \includegraphics[width=\textwidth]{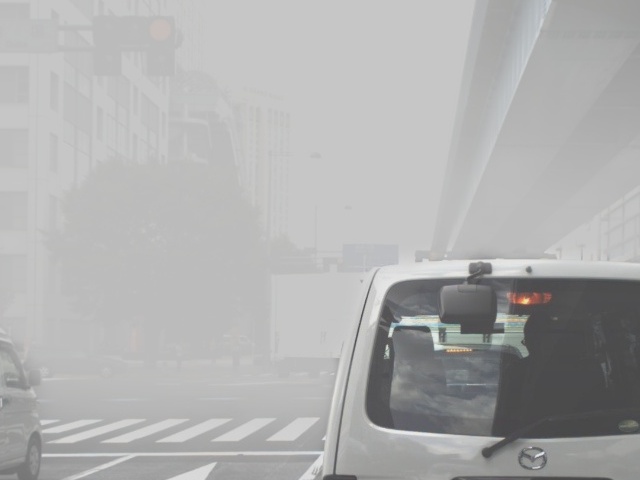}
    \end{subfigure}
    \begin{subfigure}{0.12\textwidth}
        \includegraphics[width=\textwidth]{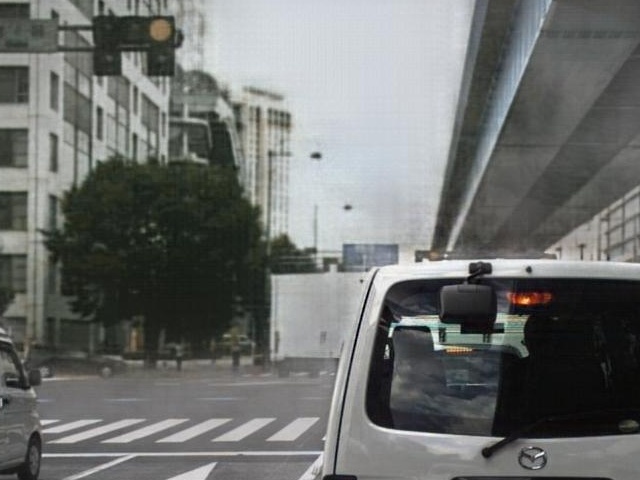}
    \end{subfigure}
    \begin{subfigure}{0.12\textwidth}
        \includegraphics[width=\textwidth]{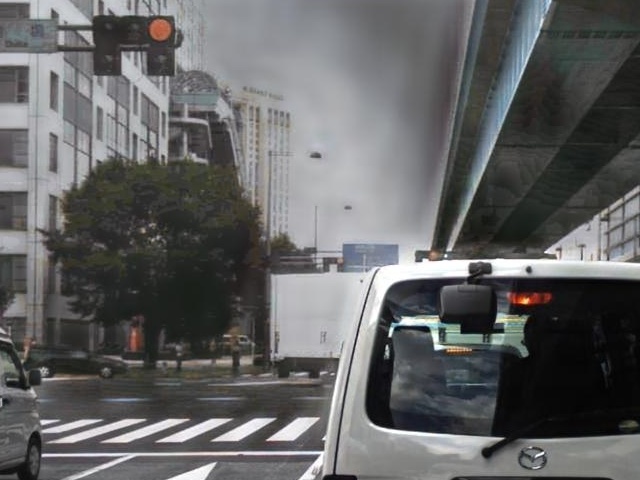}
    \end{subfigure}
    \begin{subfigure}{0.12\textwidth}
        \includegraphics[width=\textwidth]{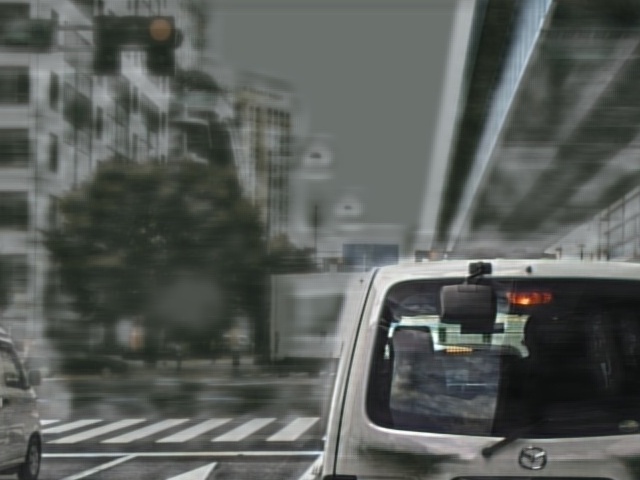}
    \end{subfigure}
    \begin{subfigure}{0.12\textwidth}
        \includegraphics[width=\textwidth]{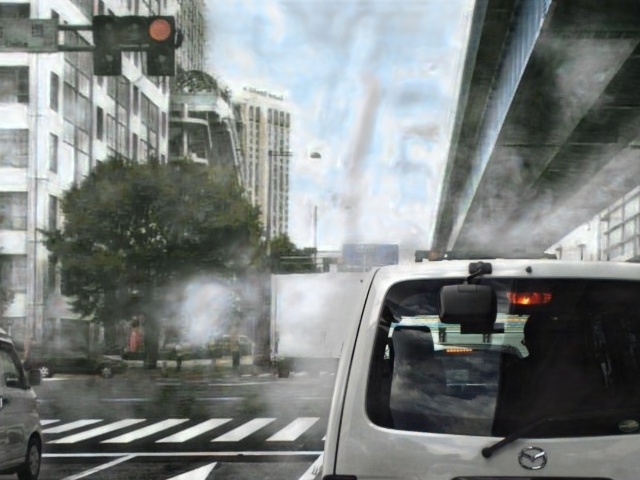}
    \end{subfigure}
    \begin{subfigure}{0.12\textwidth}
        \includegraphics[width=\textwidth]{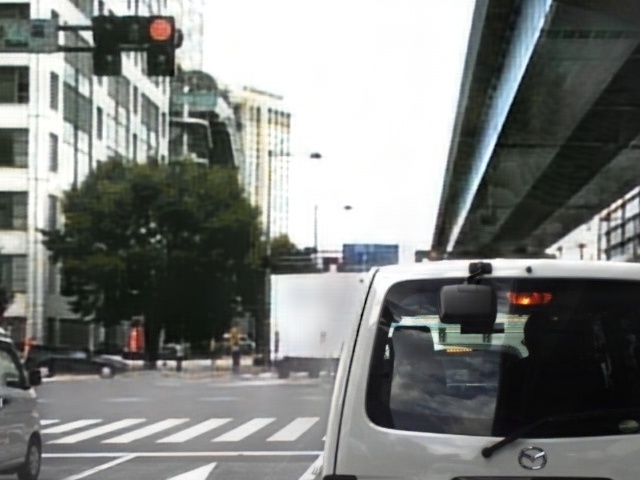}
    \end{subfigure}
    \begin{subfigure}{0.12\textwidth}
        \includegraphics[width=\textwidth]{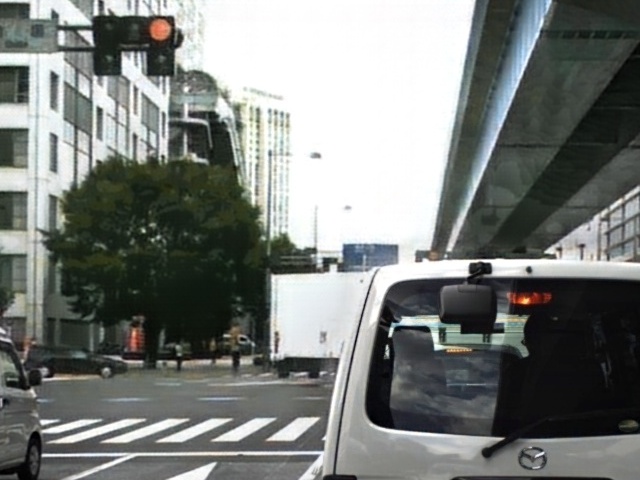}
    \end{subfigure}
    \begin{subfigure}{0.12\textwidth}
        \includegraphics[width=\textwidth]{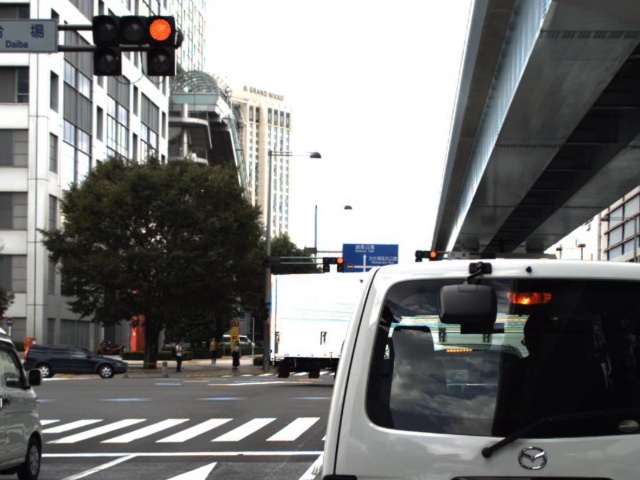}
    \end{subfigure}
    \\
    %
    %
    \begin{subfigure}{0.12\textwidth}
        \includegraphics[width=\textwidth]{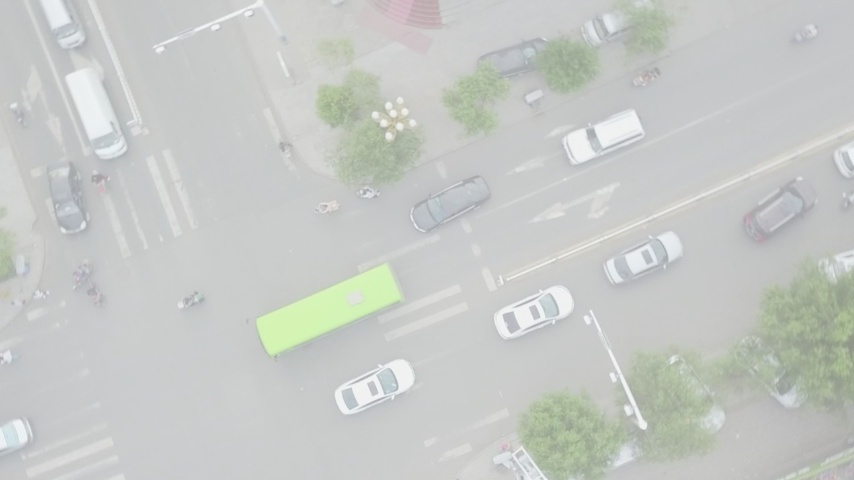}
        \caption{Hazy}
    \end{subfigure}
    \begin{subfigure}{0.12\textwidth}
        \includegraphics[width=\textwidth]{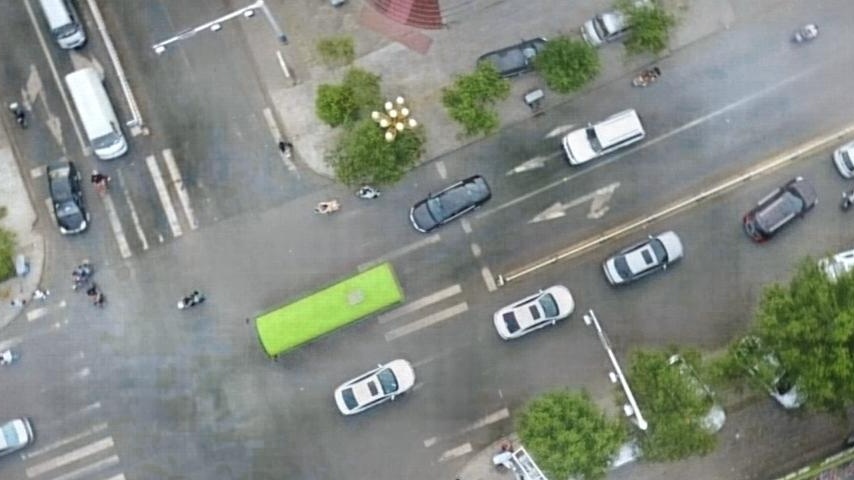}
        \caption{MSBDN~\cite{dong2020multi}}
    \end{subfigure}
    \begin{subfigure}{0.12\textwidth}
        \includegraphics[width=\textwidth]{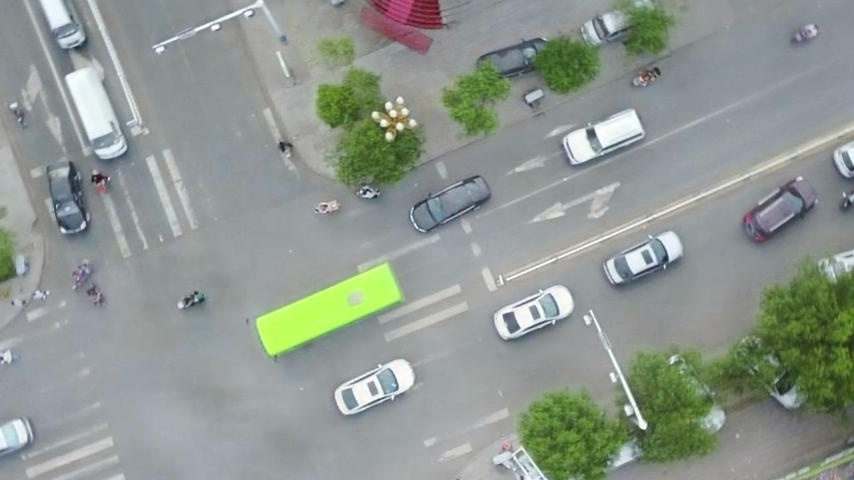}
        \caption{Dehamer~\cite{guo2022image}}
    \end{subfigure}
    \begin{subfigure}{0.12\textwidth}
        \includegraphics[width=\textwidth]{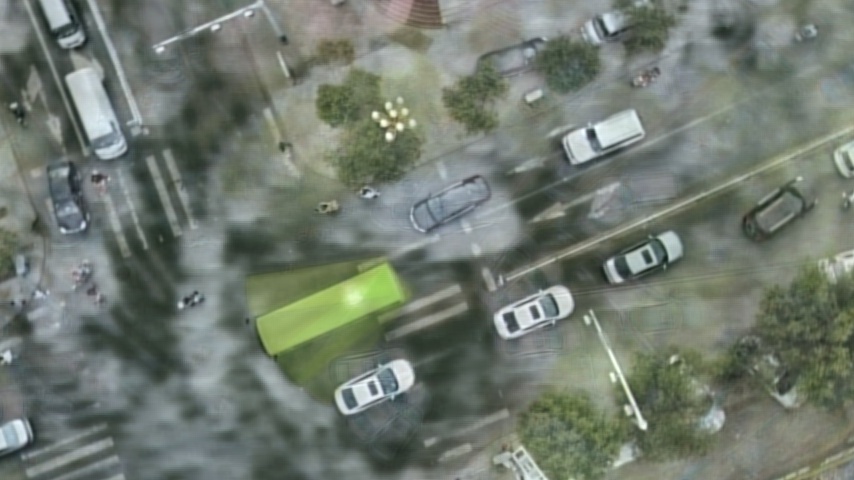}
        \caption{VDH~\cite{ren2018deep}}
    \end{subfigure}
    \begin{subfigure}{0.12\textwidth}
        \includegraphics[width=\textwidth]{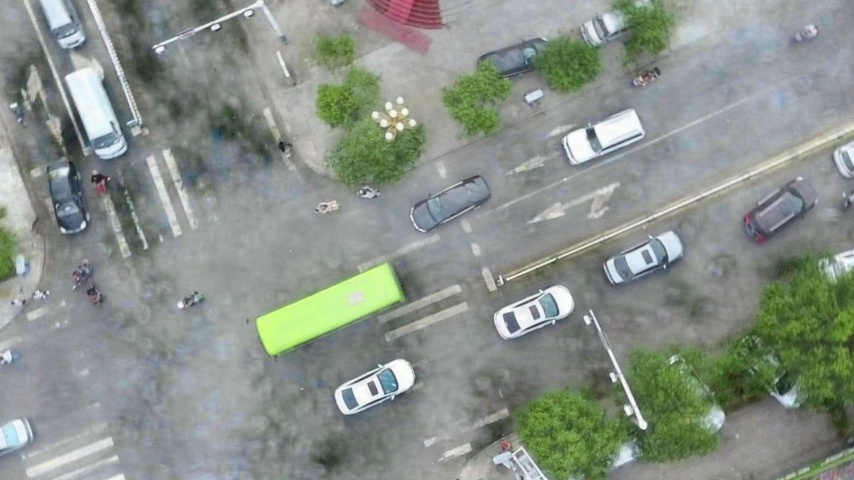}
        \caption{EDVR~\cite{wang2019edvr}}
    \end{subfigure}
    \begin{subfigure}{0.12\textwidth}
        \includegraphics[width=\textwidth]{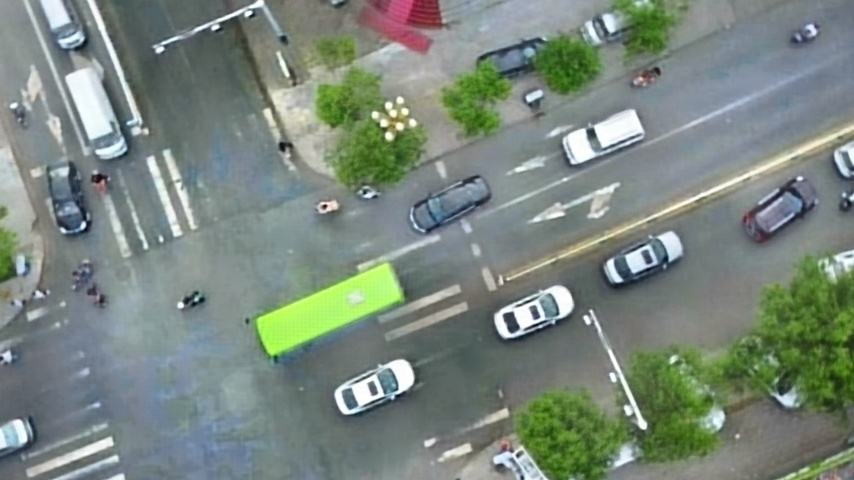}
        \caption{BasicVSR++~\cite{chan2021basicvsr++}}
    \end{subfigure}
    \begin{subfigure}{0.12\textwidth}
        \includegraphics[width=\textwidth]{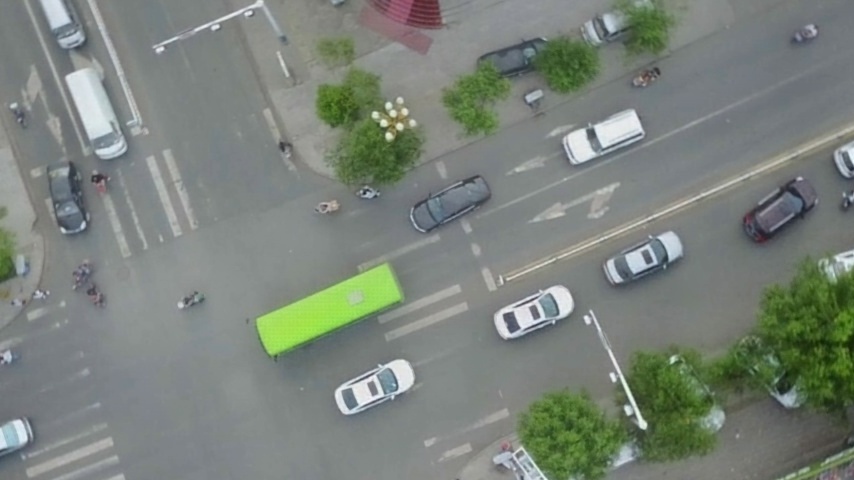}
        \caption{\textbf{Our method}}
    \end{subfigure}
    \begin{subfigure}{0.12\textwidth}
        \includegraphics[width=\textwidth]{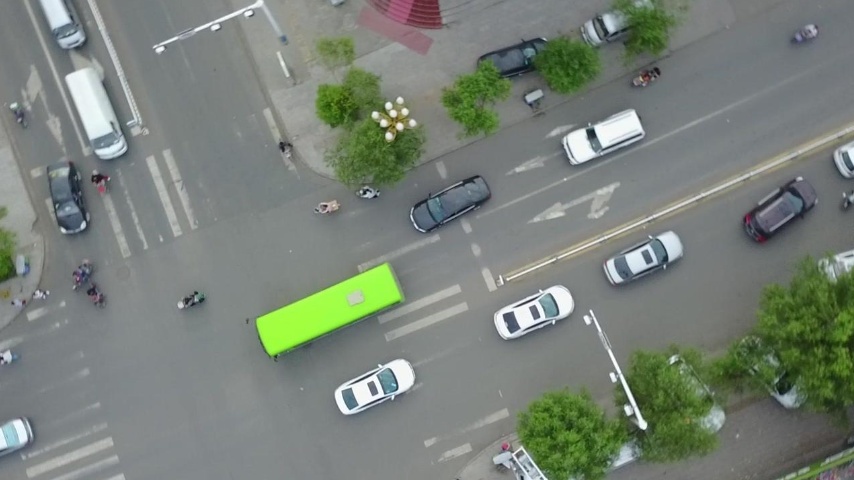}
        \caption{GT}
    \end{subfigure}
    \\
    \vspace{-2mm}
    \caption{\textbf{Visual results on our HazeWorld.} Our method clearly removes haze and keeps more details. ``GT'' denotes the ground truth.}
    \label{fig:visual_ours}
    \vspace{-2mm}
\end{figure*}

\begin{figure*}
    \centering
    \captionsetup[subfigure]{labelformat=empty,justification=centering}
    \begin{subfigure}{0.12\textwidth}
        \includegraphics[width=\textwidth]{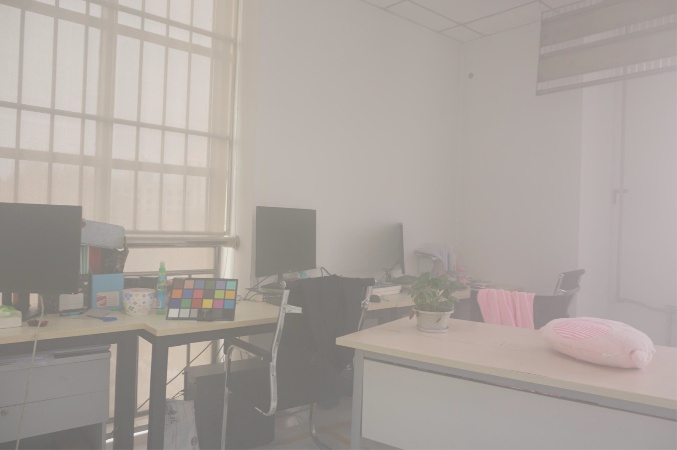}
        \caption{15.40 dB}
    \end{subfigure}
    \begin{subfigure}{0.12\textwidth}
        \includegraphics[width=\textwidth]{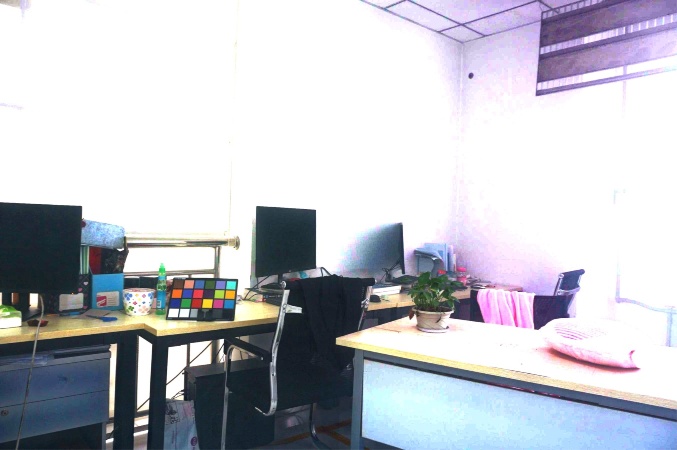}
        \caption{9.93 dB}
    \end{subfigure}
    \begin{subfigure}{0.12\textwidth}
        \includegraphics[width=\textwidth]{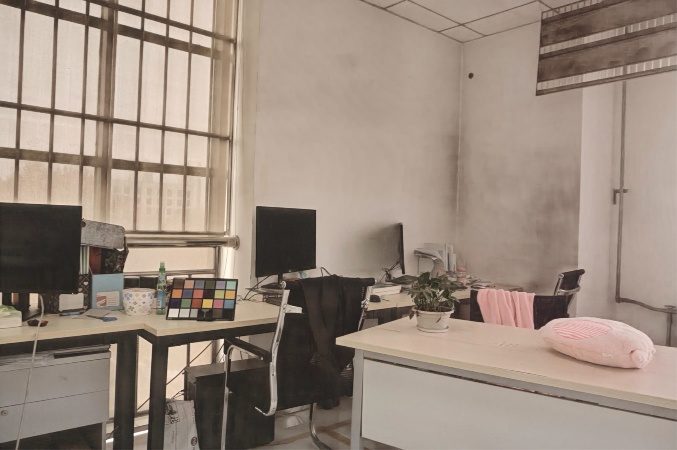}
        \caption{22.08 dB}
    \end{subfigure}
    \begin{subfigure}{0.12\textwidth}
        \includegraphics[width=\textwidth]{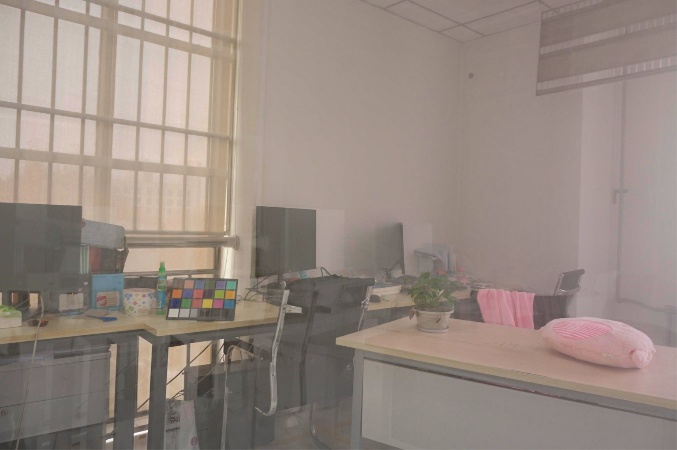}
        \caption{19.88 dB}
    \end{subfigure}
    \begin{subfigure}{0.12\textwidth}
        \includegraphics[width=\textwidth]{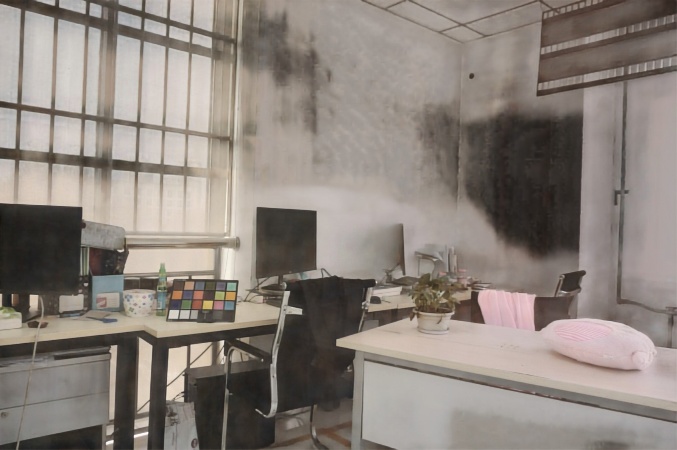}
        \caption{17.95 dB}
    \end{subfigure}
    \begin{subfigure}{0.12\textwidth}
        \includegraphics[width=\textwidth]{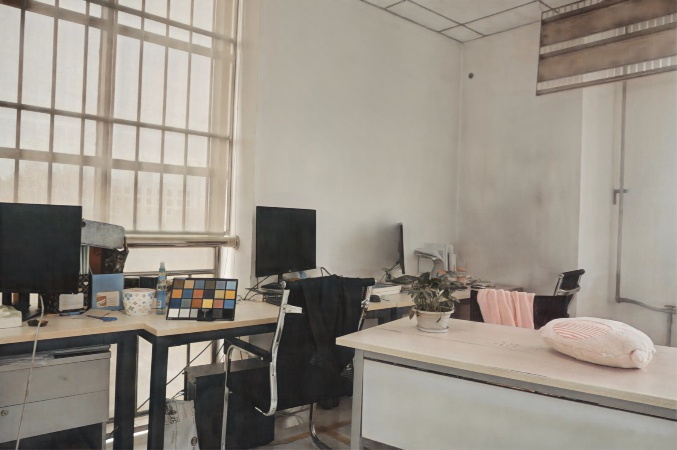}
        \caption{23.37 dB}
    \end{subfigure}
    \begin{subfigure}{0.12\textwidth}
        \includegraphics[width=\textwidth]{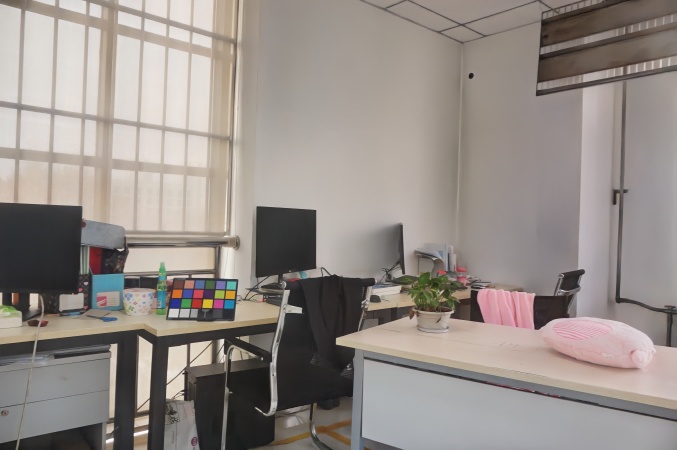}
        \caption{24.63 dB}
    \end{subfigure}
    \begin{subfigure}{0.12\textwidth}
        \includegraphics[width=\textwidth]{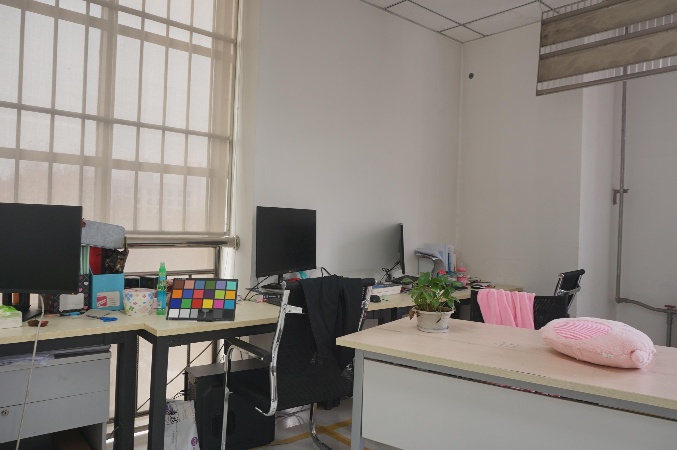}
        \caption{PSNR}
    \end{subfigure}
    \\
    \begin{subfigure}{0.12\textwidth}
        \includegraphics[width=\textwidth]{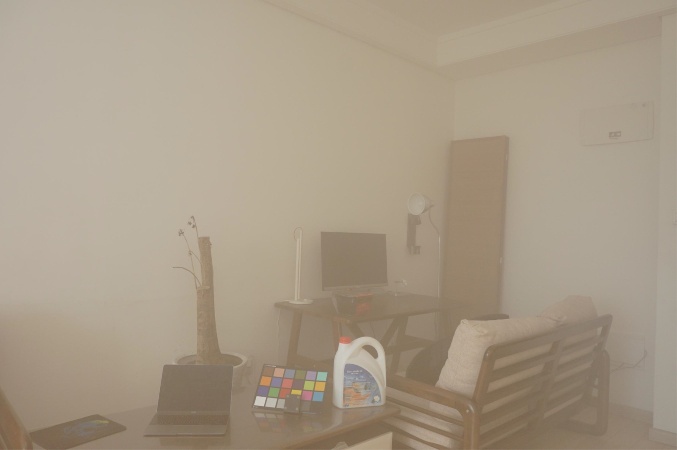}
        \caption{19.19 dB\\Hazy Input}
    \end{subfigure}
    \begin{subfigure}{0.12\textwidth}
        \includegraphics[width=\textwidth]{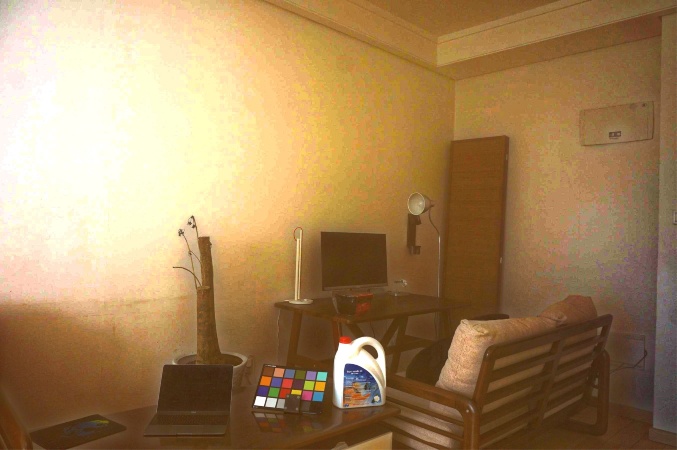}
        \caption{13.25 dB\\DCP~\cite{he2010single}}
    \end{subfigure}
    \begin{subfigure}{0.12\textwidth}
        \includegraphics[width=\textwidth]{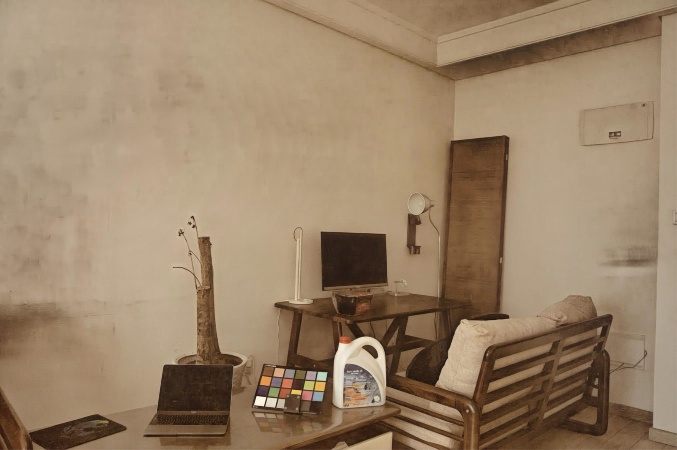}
        \caption{20.62 dB\\MSBDN~\cite{dong2020multi}}
    \end{subfigure}
    \begin{subfigure}{0.12\textwidth}
        \includegraphics[width=\textwidth]{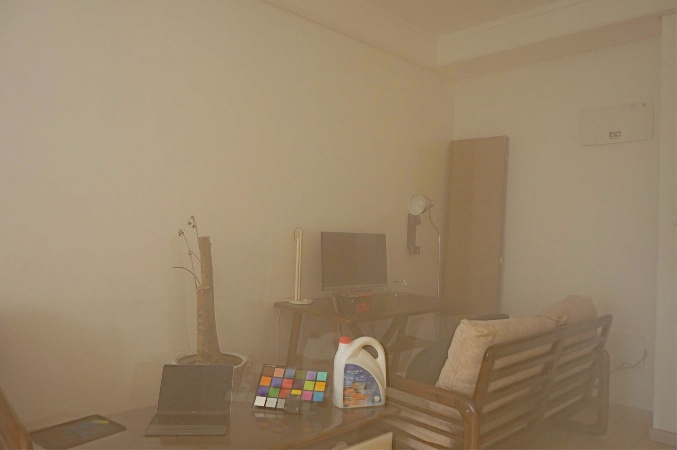}
        \caption{18.07 dB\\VDH~\cite{ren2018deep}}
    \end{subfigure}
    \begin{subfigure}{0.12\textwidth}
        \includegraphics[width=\textwidth]{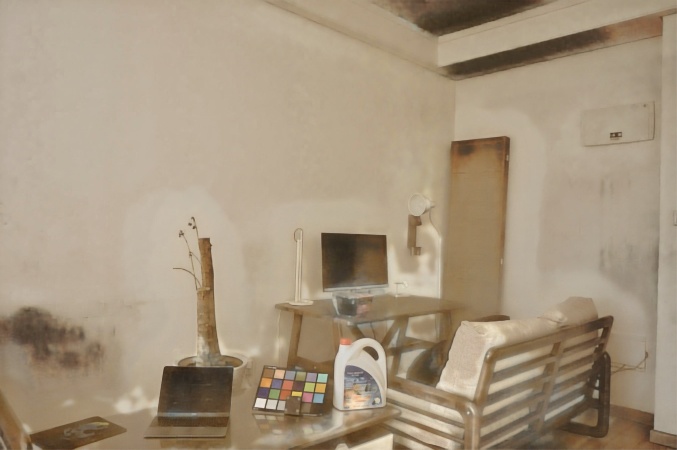}
        \caption{18.91 dB \\ EDVR~\cite{wang2019edvr}}
    \end{subfigure}
    \begin{subfigure}{0.12\textwidth}
        \includegraphics[width=\textwidth]{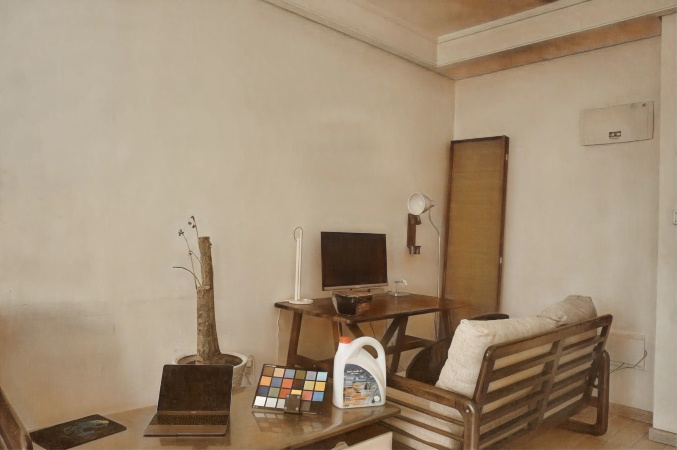}
        \caption{24.77 dB \\ CG-IDN~\cite{zhang2021learning}}
    \end{subfigure}
    \begin{subfigure}{0.12\textwidth}
        \includegraphics[width=\textwidth]{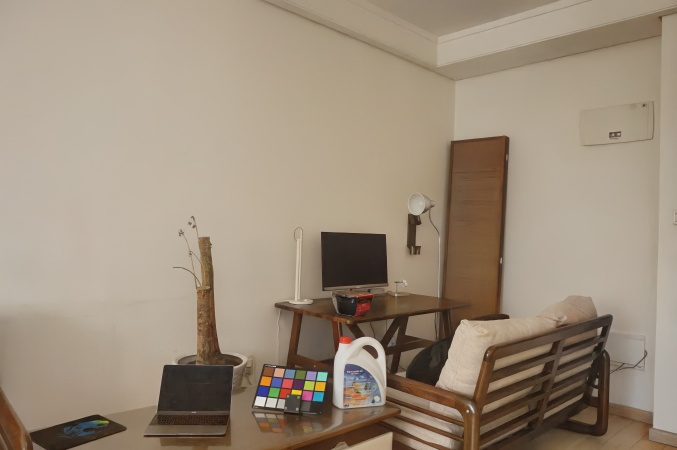}
        \caption{26.33 dB \\ \textbf{Our method}}
    \end{subfigure}
    \begin{subfigure}{0.12\textwidth}
        \includegraphics[width=\textwidth]{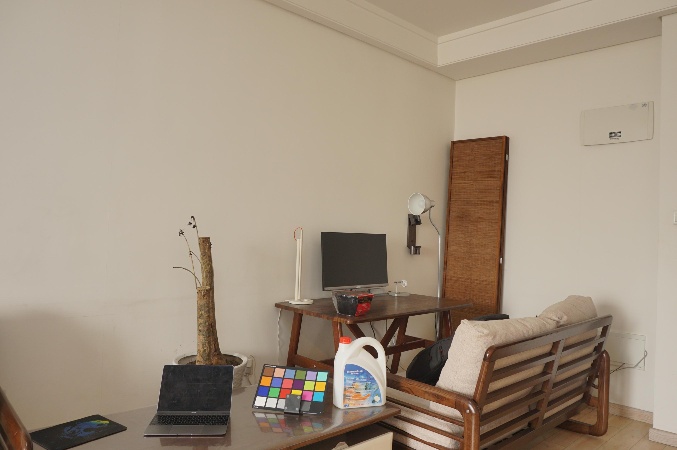}
        \caption{PSNR \\ GT}
    \end{subfigure}
    \\
    \vspace{-2mm}
    \caption{\textbf{Visual results on REVIDE~\cite{zhang2021learning}.}
    Our method produces dehazed results with less haze artifact and color distortion.
    }
    \label{fig:visual_revide}
    \vspace{-5mm}
\end{figure*}

\subsection{Settings}
\noindent
\textbf{Datasets.}
We evaluate the effectiveness of the proposed MAP-Net on our dataset, \ie, HazeWorld, and the widely-used REVIDE dataset~\cite{zhang2021learning}.
%
HazeWorld contains 3,588 training videos and 1,496 testing videos.
%
Meanwhile, REVIDE consists of 42 training videos and 6 testing videos.

\vspace{2mm}
\noindent
\textbf{Evaluation metrics.}
We utilize PSNR and SSIM to quantitatively evaluate the dehazing performance.

\vspace{2mm}
\noindent
\textbf{Comparison methods.}
On HazeWorld, we compare our method against state-of-the-art methods, including ten image dehazing methods (\ie, DCP~\cite{he2010single}, AOD~\cite{li2017aod}, GDN~\cite{liu2019griddehazenet}, DM2F~\cite{deng2019deep}, FFA~\cite{qin2020ffa}, MSBDN~\cite{dong2020multi}, UHD~\cite{zheng2021ultra}, AECR~\cite{wu2021contrastive}, Dehamer~\cite{guo2022image}, and DehazeFormer~\cite{song2023vision}), and three video dehazing methods (\ie, EVD~\cite{li2018end}, VDH~\cite{ren2018deep}, and CG-IDN~\cite{lai2018learning}).
We also compare several video restoration methods, including FastDVD~\cite{tassano2020fastdvdnet}, EDVR~\cite{wang2019edvr}, NCFL~\cite{huang2022neural}, and BasicVSR++~\cite{chan2021basicvsr++}.
On REVIDE, we compare MAP-Net with state-of-the-art methods of~\cite{zhang2021learning}. 

\vspace{2mm}
\noindent
\textbf{Implementation details.}
We use the AdamW optimizer and the polynomial scheduler.
The initial learning rate is set as 2$\times$10${}^{-4}$.
The total number of iterations is 40K.
The batch size is eight, and the patch size of input video frames is 256$\times$256.
The weights ${\lambda}_{phy}$ and ${\lambda}_{flow}$ in \cref{eq:loss} are empirically set as 0.2 and 0.04.
%

\subsection{Comparisons with State-of-the-Art Methods}
\noindent
\textbf{Quantitative comparison.}
Table~\ref{tab:exp-hazeworld} summarizes the quantitative results of our network and compared methods on HazeWorld.
%
%
From these quantitative results, we can find that our method outperforms other baselines by a significant margin.
Specifically, among all compared methods, BasicVSR++~\cite{chan2021basicvsr++} and DehazeFormer~\cite{song2023vision} achieve the best PSNR score of 26.06 and the best SSIM score of 0.9286, respectively.
More importantly, our MAP-Net has a PSNR improvement of 1.06~dB over BasicVSR++, while our method has an SSIM gain of 0.0063 over DehazeFormer.
%
%

%
Table~\ref{tab:exp-REVIDE} shows the PSNR and SSIM of our network and state-of-the-art methods on REVIDE.
Among all compared methods, NCFL~\cite{huang2022neural} has the best PSNR (23.63~dB) and the best SSIM (0.8925).
And our method further improves the PSNR from 23.63~dB to 24.16~dB and the SSIM from 0.8925 to 0.9043.


\begin{figure*}
    \centering
    \captionsetup[subfigure]{labelformat=empty,justification=centering}
    \begin{subfigure}{0.12\hsize}
        \includegraphics[width=\columnwidth]{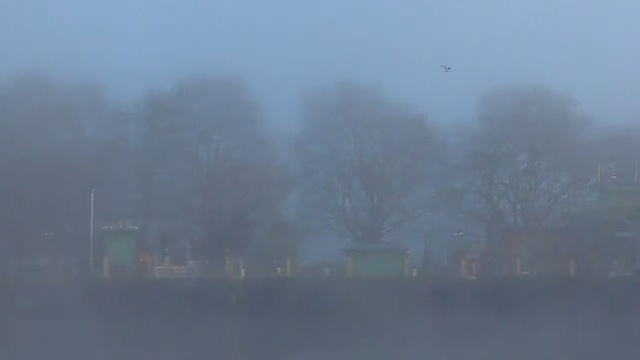}
    \end{subfigure}
    \begin{subfigure}{0.12\hsize}
        \includegraphics[width=\columnwidth]{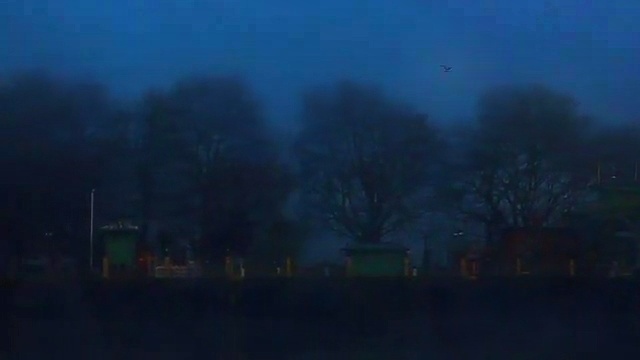}
    \end{subfigure}
    \begin{subfigure}{0.12\hsize}
        \includegraphics[width=\hsize]{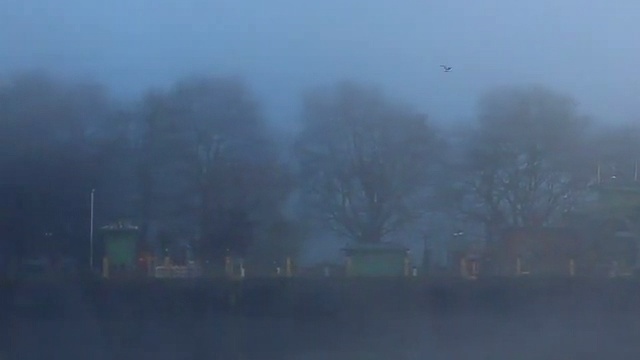}
    \end{subfigure}
    \begin{subfigure}{0.12\hsize}
        \includegraphics[width=\hsize]{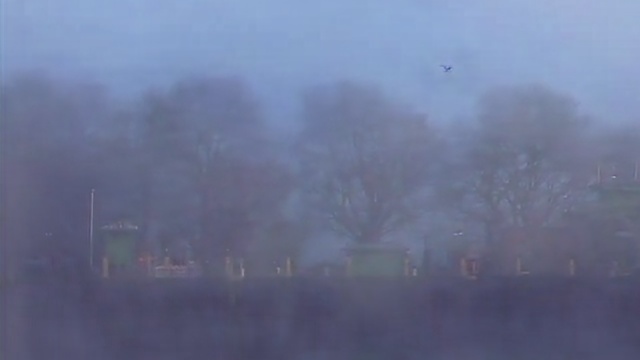}
    \end{subfigure}
    \begin{subfigure}{0.12\hsize}
        \includegraphics[width=\hsize]{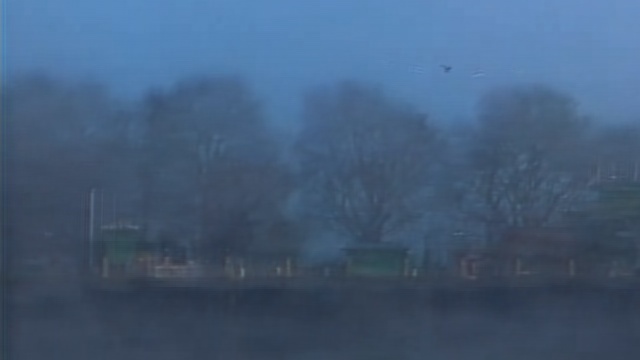}
    \end{subfigure}
    \begin{subfigure}{0.12\hsize}
        \includegraphics[width=\hsize]{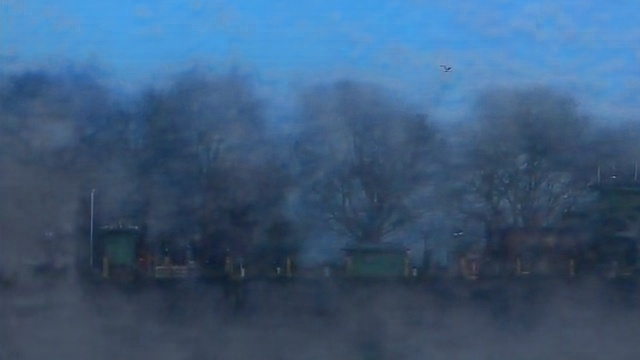}
    \end{subfigure}
    \begin{subfigure}{0.12\hsize}
        \includegraphics[width=\hsize]{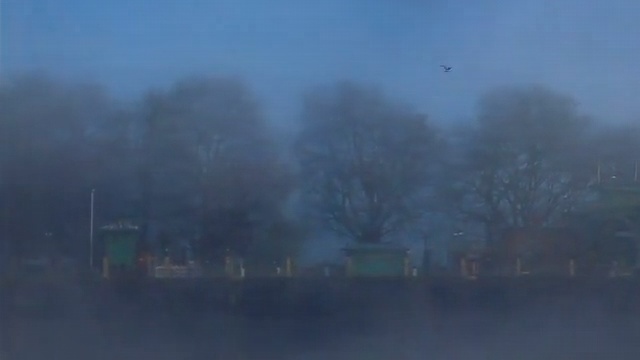}
    \end{subfigure}
    \begin{subfigure}{0.12\hsize}
        \includegraphics[width=\hsize]{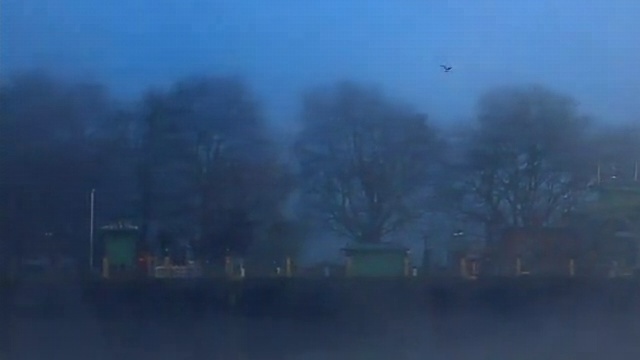}
    \end{subfigure}
    \\
    \begin{subfigure}{0.12\hsize}
        \includegraphics[width=\columnwidth]{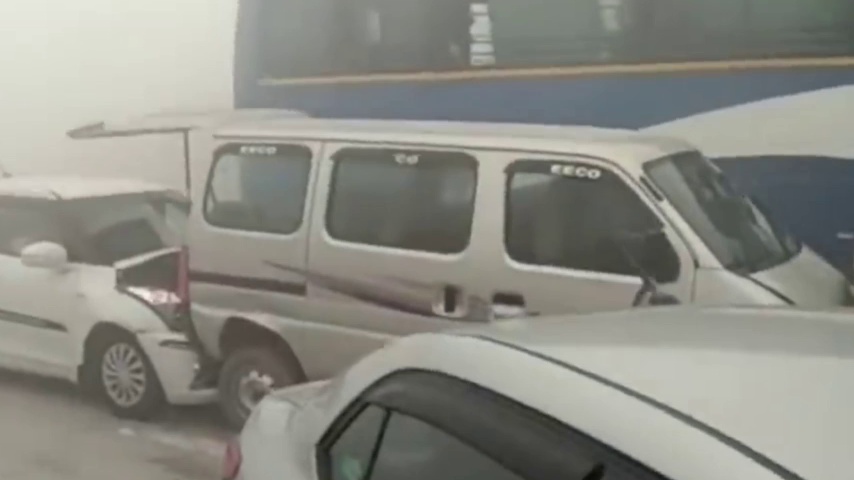}
    \end{subfigure}
    \begin{subfigure}{0.12\hsize}
        \includegraphics[width=\columnwidth]{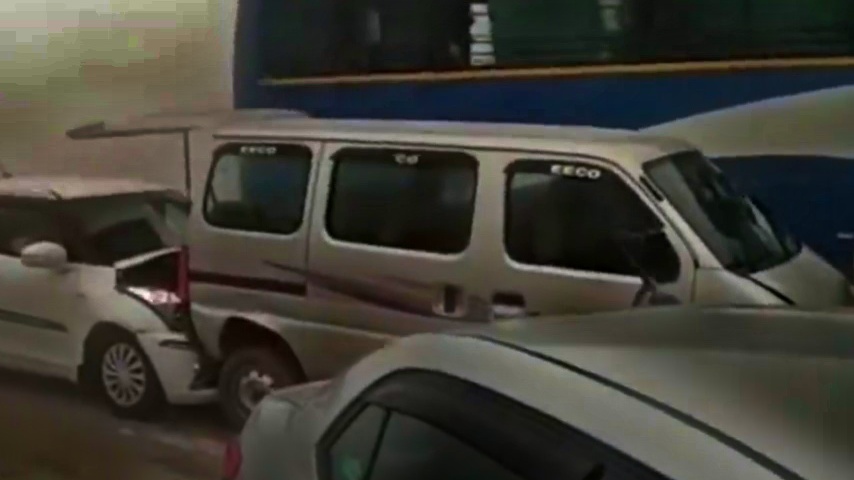}
    \end{subfigure}
    \begin{subfigure}{0.12\hsize}
        \includegraphics[width=\hsize]{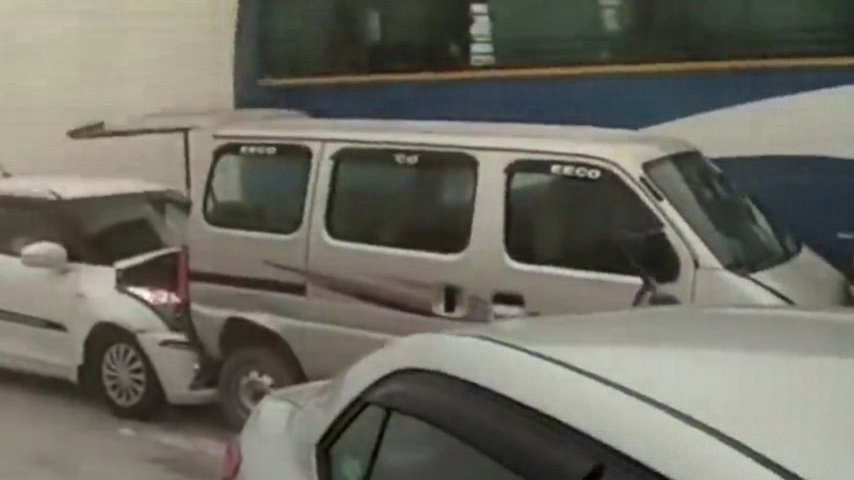}
    \end{subfigure}
    \begin{subfigure}{0.12\hsize}
        \includegraphics[width=\hsize]{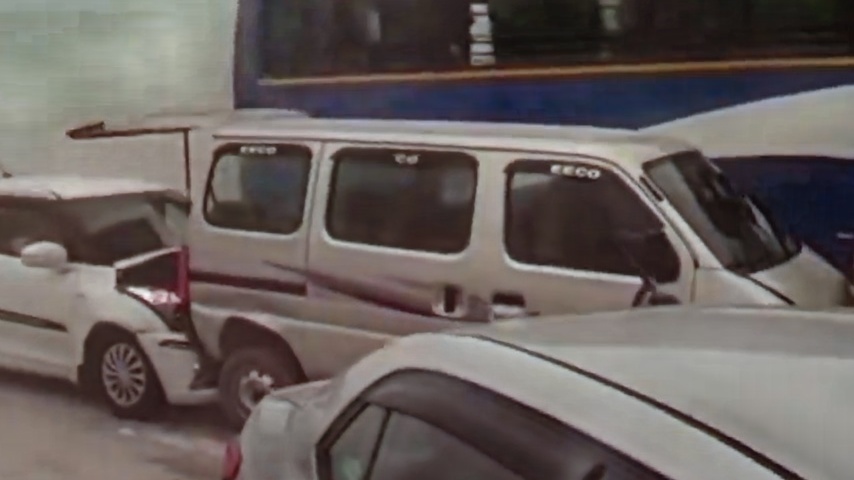}
    \end{subfigure}
    \begin{subfigure}{0.12\hsize}
        \includegraphics[width=\hsize]{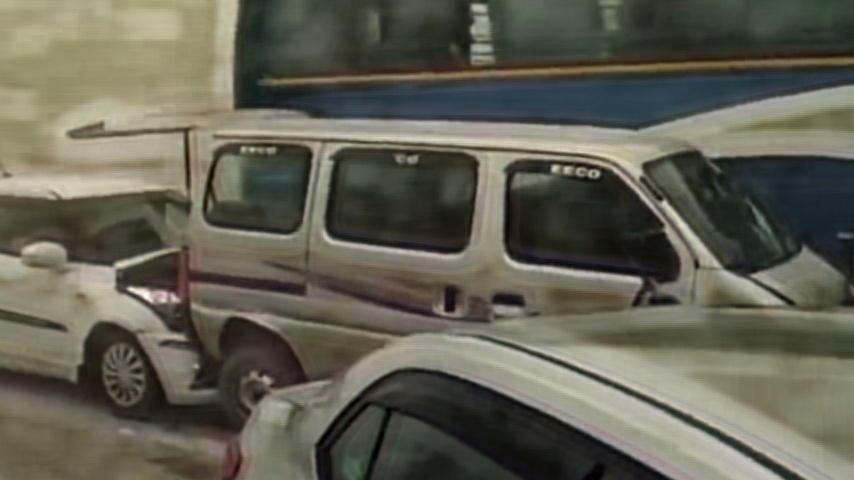}
    \end{subfigure}
    \begin{subfigure}{0.12\hsize}
        \includegraphics[width=\hsize]{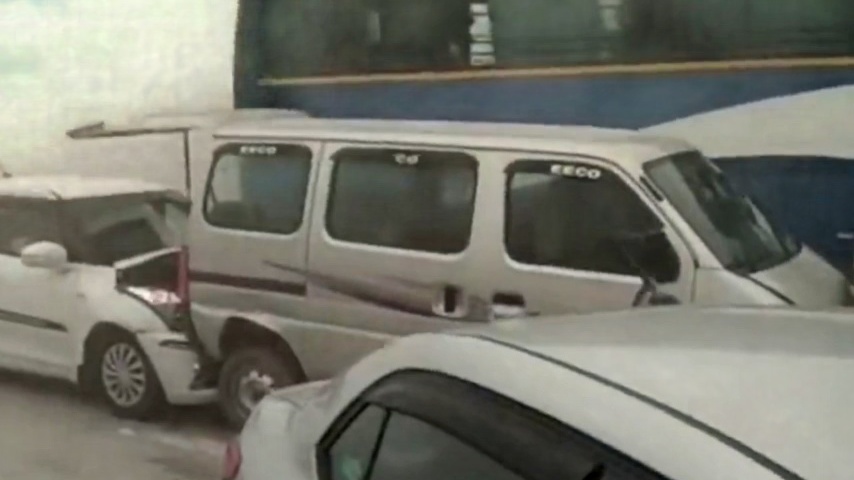}
    \end{subfigure}
    \begin{subfigure}{0.12\hsize}
        \includegraphics[width=\hsize]{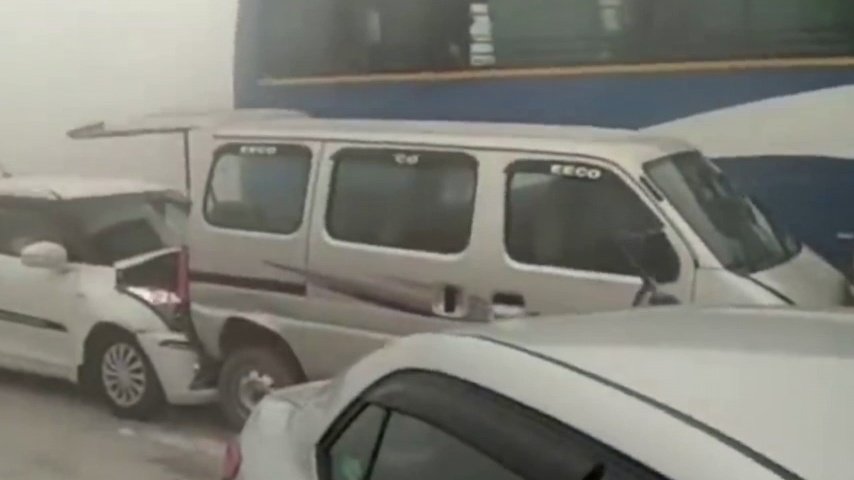}
    \end{subfigure}
    \begin{subfigure}{0.12\hsize}
        \includegraphics[width=\hsize]{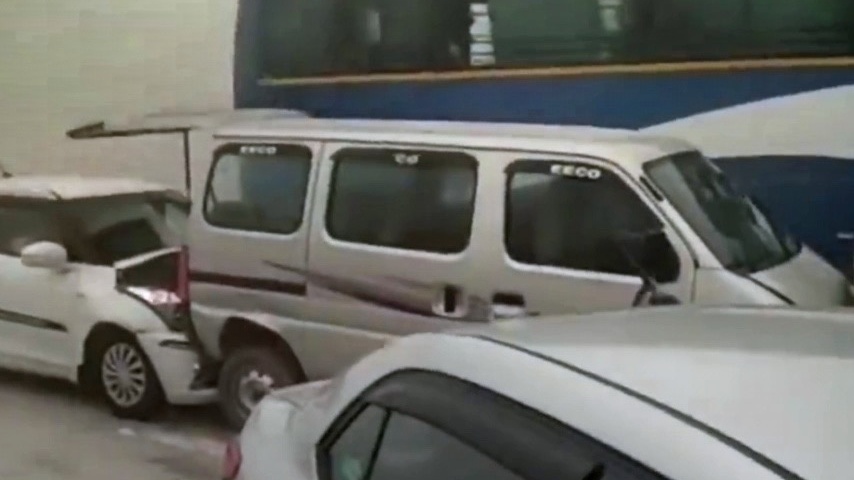}
    \end{subfigure}
    \\
    \begin{subfigure}{0.12\hsize}
        \includegraphics[width=\columnwidth]{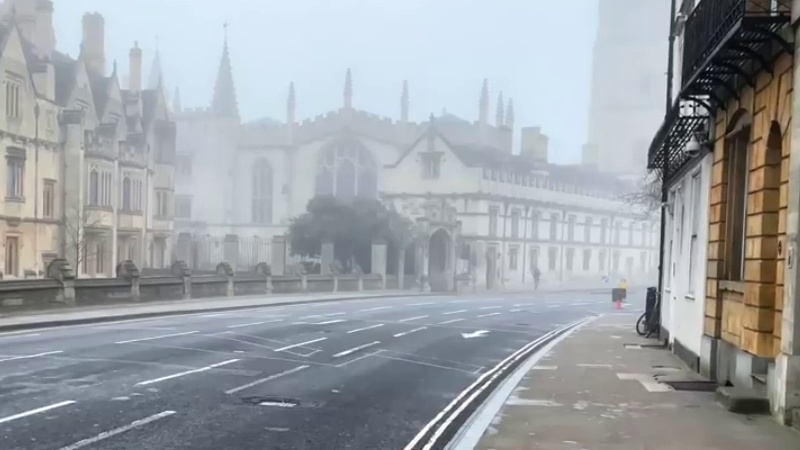}
    \end{subfigure}
    \begin{subfigure}{0.12\hsize}
        \includegraphics[width=\columnwidth]{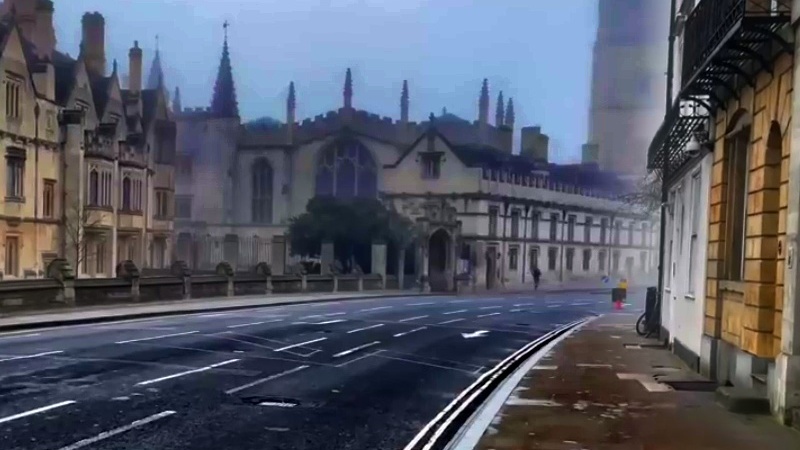}
    \end{subfigure}
    \begin{subfigure}{0.12\hsize}
        \includegraphics[width=\hsize]{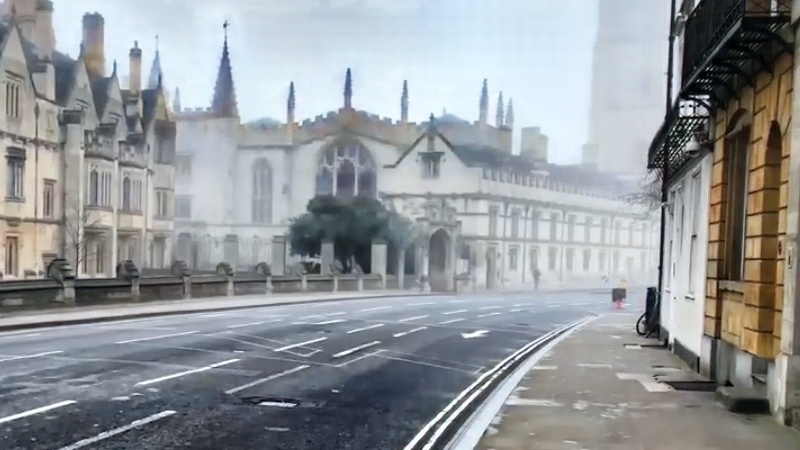}
    \end{subfigure}
    \begin{subfigure}{0.12\hsize}
        \includegraphics[width=\hsize]{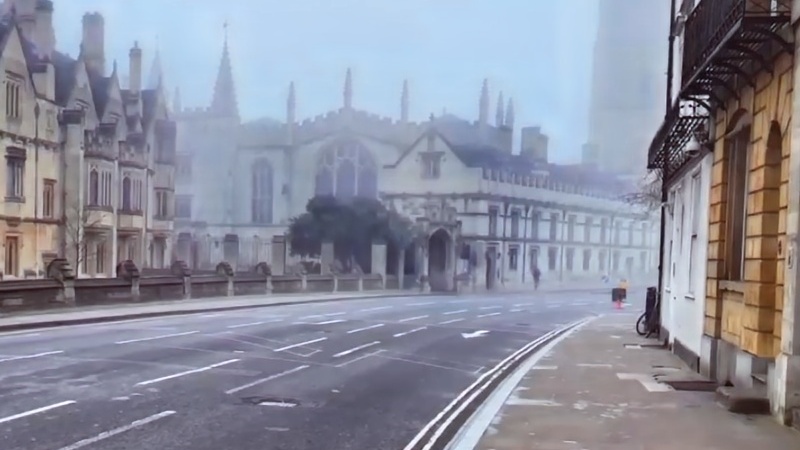}
    \end{subfigure}
    \begin{subfigure}{0.12\hsize}
        \includegraphics[width=\hsize]{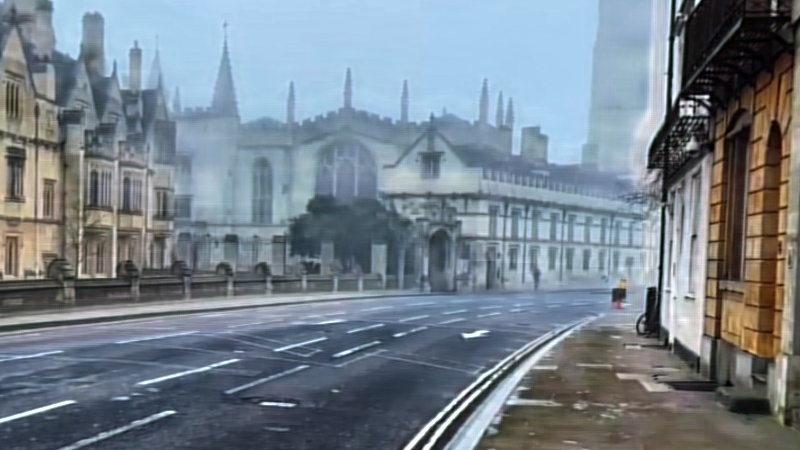}
    \end{subfigure}
    \begin{subfigure}{0.12\hsize}
        \includegraphics[width=\hsize]{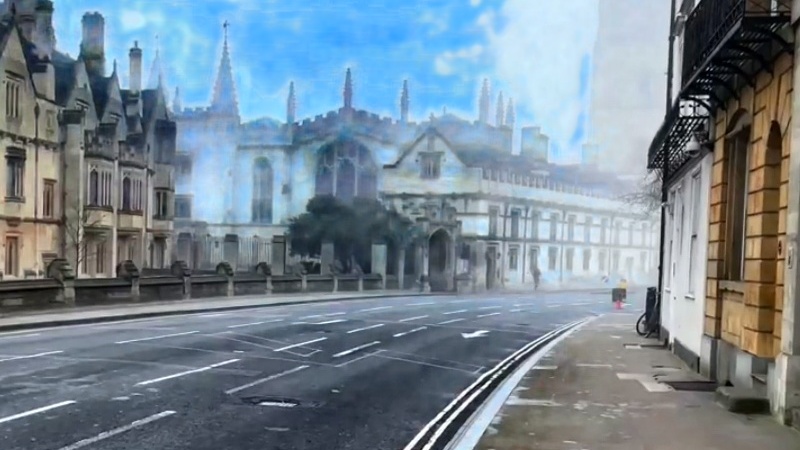}
    \end{subfigure}
    \begin{subfigure}{0.12\hsize}
        \includegraphics[width=\hsize]{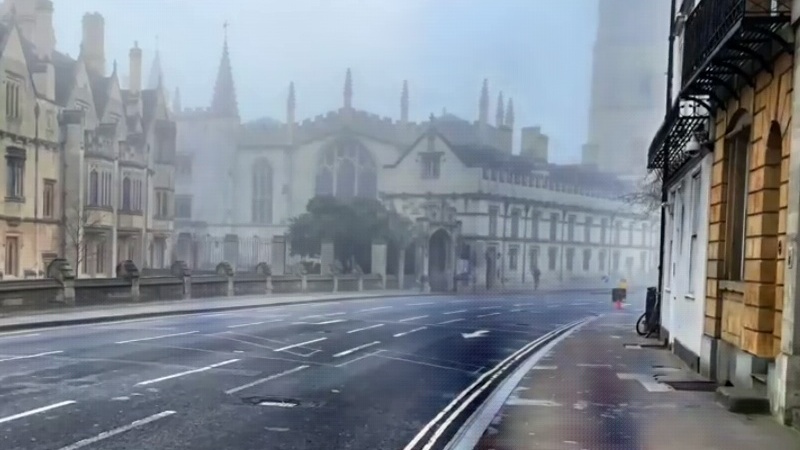}
    \end{subfigure}
    \begin{subfigure}{0.12\hsize}
        \includegraphics[width=\hsize]{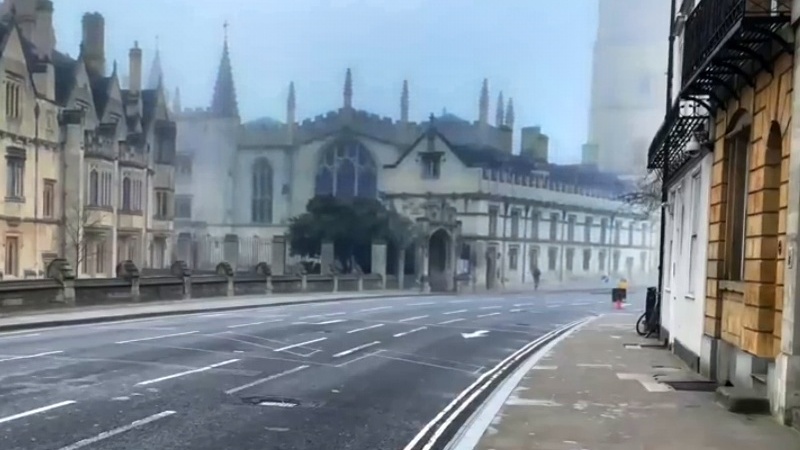}
    \end{subfigure}
    \\
    \begin{subfigure}{0.12\hsize}
        \includegraphics[width=\columnwidth]{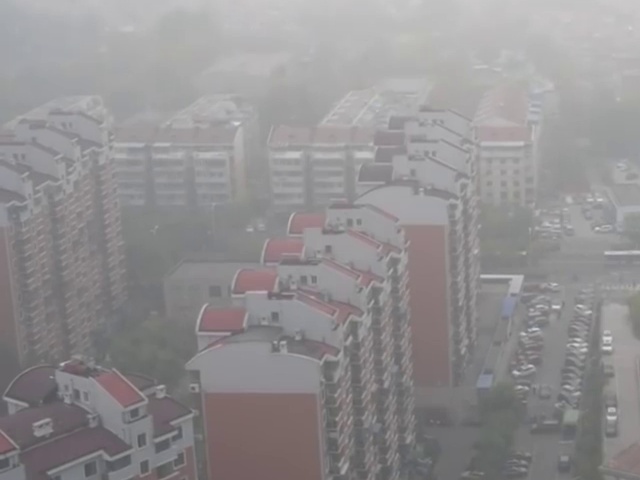}
        \caption{Hazy}
    \end{subfigure}
    \begin{subfigure}{0.12\hsize}
        \includegraphics[width=\columnwidth]{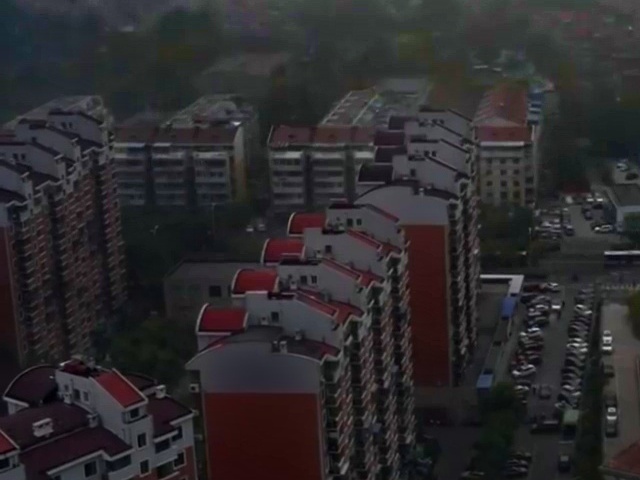}
        \caption{DCP~\cite{he2010single}}
    \end{subfigure}
    \begin{subfigure}{0.12\hsize}
        \includegraphics[width=\hsize]{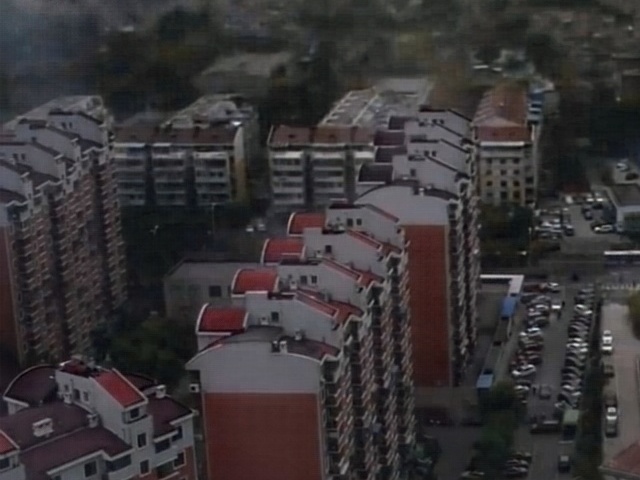}
        \caption{MSBDN~\cite{dong2020multi}}
    \end{subfigure}
    \begin{subfigure}{0.12\hsize}
        \includegraphics[width=\hsize]{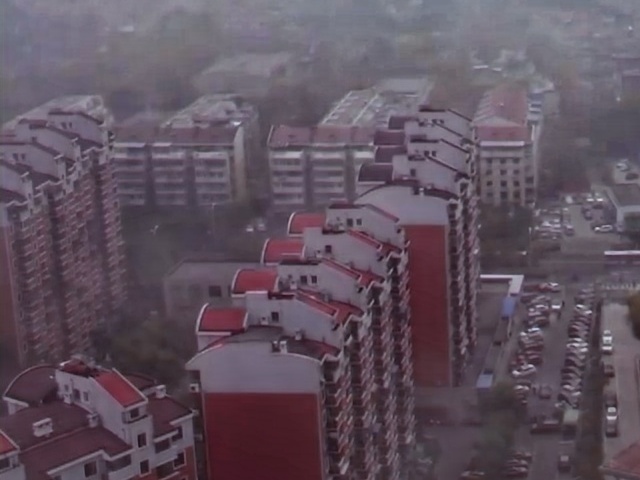}
        \caption{Dehamer~\cite{guo2022image}}
    \end{subfigure}
    \begin{subfigure}{0.12\hsize}
        \includegraphics[width=\hsize]{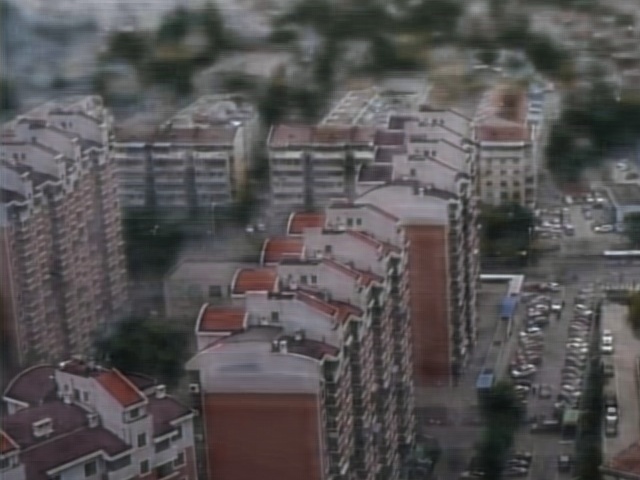}
        \caption{VDH~\cite{ren2018deep}}
    \end{subfigure}
    \begin{subfigure}{0.12\hsize}
        \includegraphics[width=\hsize]{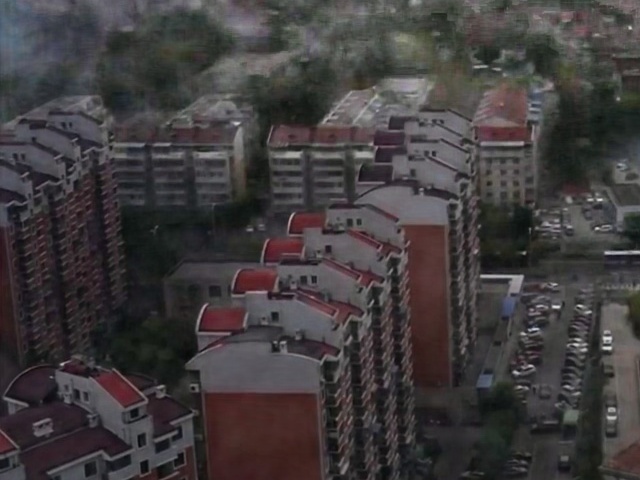}
        \caption{EDVR~\cite{wang2019edvr}}
    \end{subfigure}
    \begin{subfigure}{0.12\hsize}
        \includegraphics[width=\hsize]{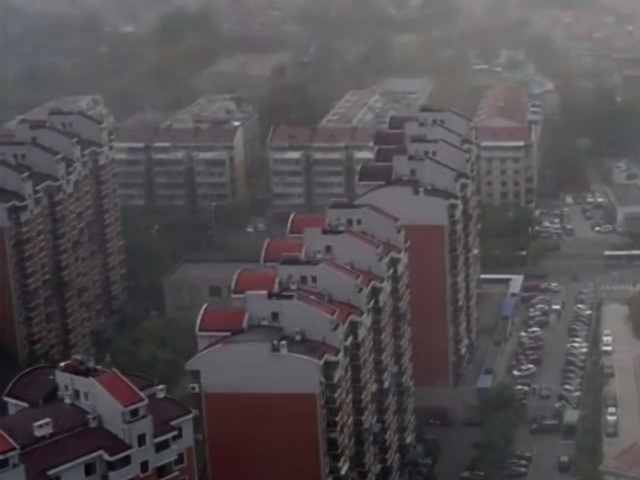}
        \caption{BasicVSR++~\cite{chan2021basicvsr++}}
    \end{subfigure}
    \begin{subfigure}{0.12\hsize}
        \includegraphics[width=\hsize]{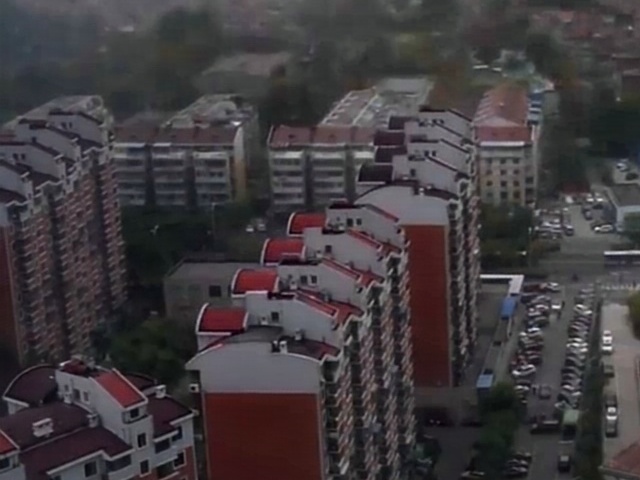}
        \caption{\textbf{Our method}}
    \end{subfigure}
    \\
    \vspace{-2mm}
    \caption{\textbf{Visual results on the real outdoor hazy videos.} Our method generates frames with more natural color and less haze remaining. 
    }
    \label{fig:visual_real}
    \vspace{-5mm}
\end{figure*}

\vspace{2mm}
\noindent
\textbf{Qualitative comparison.}
\cref{fig:visual_ours} and \cref{fig:visual_revide} visually compare dehazed results produced by our network and state-of-the-art methods on video frames from HazeWorld and REVIDE.
Apparently, compared methods often tend to introduce color distortion, darken several areas, or preserve some haze in their dehazed results, while our MAP-Net can effectively remove haze, avoid color distortion, and better recover the underlying clear frames. 
And the predicted haze-free results produced by our method are closest to ground truths shown in the last column of \cref{fig:visual_ours} and \cref{fig:visual_revide}.

Moreover, we also compare our network against state-of-the-art methods on real-world hazy videos, and the results are shown in~\cref{fig:visual_real}.
From these visual results, we can find that existing methods tend to darken many areas, or maintain some haze.
Compared to these methods, our network has a higher visual quality and less color distortion; see the last column of~\cref{fig:visual_real}.
%

%
\begin{table}[t]
\caption{\textbf{Comparison of downstream effectiveness.} VPQ, mIoU, RMSE, and J\&F are metrics for panoptic segmentation, semantic segmentation, depth estimation, and object segmentation.}
\label{tab:downstream}
\vspace{-1mm}
\centering
\begin{adjustbox}{width=\linewidth}
\begin{tabular}{lccccccc}
\hline
Method                     & Hazy   & MSBDN~\cite{dong2020multi} & Dehamer~\cite{guo2022image} & EDVR~\cite{wang2019edvr} & BasicVSR++~\cite{chan2021basicvsr++} & Ours            & GT     \\
\hline
Cityscapes-VPQ $\uparrow$  & 21.7   & 40.2                       & 43.4                        & 47.2                     & 45.9                                 & \textbf{48.5}   & 56.5   \\
Cityscapes-mIoU $\uparrow$ & 51.8   & 47.0                       & 54.1                        & 64.8                     & 63.6                                 & \textbf{66.2}   & 75.4   \\
DDAD-RMSE $\downarrow$     & 21.21  & 14.99                      & 15.01                       & 15.26                    & 14.89                                & \textbf{14.71 } & 14.36 \\
DAVIS-J\&F $\uparrow$      & 76.3   & 79.2                       & 79.4                        & 79.3                     & 79.4                                 & \textbf{80.0}   & 81.3   \\
\hline
\end{tabular}
\end{adjustbox}
\vspace{-5mm}
\end{table}

\vspace{2mm}
\noindent
\textbf{Applications.}
Our video dehazing method benefits several downstream applications, including video panoptic segmentation~\cite{kim2020video}, object segmentation~\cite{pont20172017}, depth estimation~\cite{guizilini20203d}, and image semantic segmentation~\cite{cordts2016cityscapes}.
%
To verify this, we choose four different methods~\cite{kim2020video,oh2019video,guizilini20203d,zhao2017pyramid} for corresponding downstream application validation, and obtain results on the input hazy videos, the dehazed videos, and the underlying haze-free videos.
%
%
%
Table~\ref{tab:downstream} reports the quantitative results.
Apparently, the dehazed videos produced by different methods improve the downstream application performance compared to the original hazy videos.
Notably, our method can better facilitate downstream applications than other representative dehazing methods.
%
%

%
\begin{table}[t]
\caption{\textbf{Ablation studies of our MPG and MSR modules.}}
\label{tab:ablation-components}
\centering
\footnotesize
\begin{tabular}{lcccc}
\hline
         & (a)        & (b)        & (c)        & Our method \\ \hline
Basic    & \checkmark & \checkmark & \checkmark & \checkmark \\
MPG      &            & \checkmark &            & \checkmark \\
MSR      &            &            & \checkmark & \checkmark \\ \hline
PSNR     & 25.37      & 26.24      & 26.38      & \textbf{27.12}      \\
SSIM     & 0.9087     & 0.9171     & 0.9292     & \textbf{0.9349}     \\ \hline
\end{tabular}
\vspace{-3mm}
\end{table}

\subsection{Ablation Studies}
We conduct a series of ablation studies on our HazeWorld dataset to analyze the effectiveness of major components of our network.
%
%

\vspace{2mm}
\noindent
\textbf{Ablation studies of two major modules.}
We start by constructing a basic model (denoted as ``Basic''), which has the only scene decoder, and STDA only considers one previous frame for temporal alignment.
Then, we gradually add two proposed modules, \ie, MPG and MSR. 
Table~\ref{tab:ablation-components} compares their results.
Compared to ``Basic'', the MPG module and the MSR module have a PSNR improvement of 0.87~dB and 1.01~dB, respectively, due to the physical model-based priors at the MPG module, and multiple temporal haze clues at the MSR module.
Moreover, combining both MPG and MSR modules together into our method can further improve our video dehazing performance. 


\vspace{2mm}
\noindent
\textbf{Effectiveness of guidance information in our MSR module.}
As shown in ~\cref{fig:overview}, our MPG module learns prior guidance features, \ie, $\mathcal{J}$ and $\mathcal{P}$, to guide the STDA block to align video frames and the GMRA block to aggregate features in our MSR module for video dehazing.
Therefore, we conduct an ablation study to evaluate the effectiveness of the guidance information for the STDA block and the GMRA block, respectively.
We achieve this by building three baseline networks: (1) we remove guidance information from both the STDA block and the GMRA block; (2) we only use the guidance at the STDA block; (3) we only utilize the guidance at the GMRA block.
Note that we keep both the STDA block and the GMRA block in our MSR module but only remove the guidance parts during the ablation.
As shown in Table~\ref{tab:ablation-prior}, our video dehazing performance is reduced if we remove the prior guidance information from either the STDA block or the GMRA block.

\begin{table}[t]
    \caption{\textbf{Effectiveness of the guidance at the STDA and GMRA blocks of the MSR module.}}
    \label{tab:ablation-prior}
    \vspace{-2mm}
    \centering
    \begin{adjustbox}{width=0.7\columnwidth}
    \footnotesize
    \begin{tabular}{lcccc}
    \hline
    STDA   &            & \checkmark &            & \checkmark     \\
    GMRA   &            &            & \checkmark & \checkmark     \\
    \hline
    PSNR   & 26.38      & 26.92      & 26.61      & \textbf{27.12}  \\
    \hline
    \end{tabular}
    \end{adjustbox}
    \vspace{-2mm}
\end{table}
\begin{table}[t]
    \caption{\textbf{Ablation studies of the number of ranges in MSR.}}
    \label{tab:ablation-msr}
    \centering
    \vspace{-2mm}

    \begin{subtable}[t]{\hsize}
    \caption{Discussion on the number of ranges.}
    \label{tab:ablation-range}
    \centering
    \begin{adjustbox}{width=0.7\hsize}
    \footnotesize
    \begin{tabular}{lcccc}
    \hline
    \#Range  & 1       & 2       & 3 (Ours)         & 4       \\ \hline
    PSNR     & 26.24   & 26.71   & \textbf{27.12}   & 26.84   \\ \hline
    \end{tabular}
    \end{adjustbox}
    \end{subtable}

    \begin{subtable}[t]{\hsize}
    \caption{Discussion on the temporal alignment.}
    \label{tab:ablation-align}
    \centering
    \begin{adjustbox}{width=0.7\hsize}
    \small
    \begin{tabular}{lcccc}
    \hline
    Manner  & 1set    & 3sets-1range  & Our method      \\ \hline
    PSNR    & 26.39   & 26.64         & \textbf{27.12}  \\ \hline
    \end{tabular}
    \end{adjustbox}
    \end{subtable}
    \vspace{-5mm}
\end{table}

\vspace{2mm}
\noindent
\textbf{Discussion on different ranges of our MSR module.}
We perform two ablation study experiments on our MSR module: one is to discuss the number of space-time ranges used in our MSR module, while another experiment is to discuss how to leverage multiple adjacent frames when the number of ranges is fixed.
Table~\ref{tab:ablation-range} reports PSNR scores of our video dehazing with different numbers of ranges.
We can observe that our PSNR is progressively improved when we increase the number of ranges from one to three. The reason is that a larger range value involves more temporal information for aligning video frames.
However, when we further increase the range value from three to four, our video dehazing performance is reduced
since the weights are shared for the STDA blocks, where a single STDA needs to tackle different ranges of temporal alignment.
Hence, a larger range may introduce difficulty in the flow estimation, and thus we empirically set the number of frames/ranges as three.

Moreover, we further discuss how to use the adjacent video frames for temporal alignment when the number of frames is fixed as three.
Here, we construct two baselines: (1) the 1st baseline (denoted as ``1set'') is to change the three-set alignment in our method to only one set, which takes all three neighboring video frames.
(2) the 2nd baseline (denoted as ``3sets-1range'') is constructed by only using one adjacent frame in each set, \ie, frame-by-frame alignment (three sets in total).
As shown in Table~\ref{tab:ablation-align}, our method has a better PSNR value than ``1set'' and ``3sets-1range'', which indicates that considering three frames at multiple ranges incurs a better video dehazing performance.
\section{Conclusion}
This work designs a video dehazing framework via a multi-range temporal alignment network with physical prior.
Two new techniques, a memory-based physical prior guidance module and a multi-range scene radiance recovery module, are formulated to effectively explore the physical haze priors and aggregate temporal information.
We construct the first large-scale benchmark dataset for outdoor video dehazing, which enables us to evaluate the dehazing performance on various application scenarios and downstream tasks.
In the end, the experimental results on both synthetic and real conditions demonstrate the superior of our framework against the recent state-of-the-art methods.

\vspace{-4mm}
\paragraph{Limitations.}
Our method might not work well for videos with extremely heavy haze, and more prior knowledge is required.
Also, though our method achieves superior performance and faster speed than many dehazing methods, it still cannot meet the real-time requirement. 

\vspace{-4mm}
\paragraph{Acknowledgment.}
This work was partially supported by the
Research Grants Council of the Hong Kong Special Administrative Region, China (Project Reference Number: 14201321 and 14201620),
the National Natural Science Foundation of China (Grant No. 61902275),
the National Key R\&D Program of China (NO.2022ZD0160100),
and Shanghai Committee of Science and Technology (Grant No. 21DZ1100100).



{\small
\bibliographystyle{ieee_fullname}
\bibliography{ref}

\begin{thebibliography}{10}\itemsep=-1pt

\bibitem{ancuti2020nh}
Codruta~O Ancuti, Cosmin Ancuti, and Radu Timofte.
\newblock Nh-haze: An image dehazing benchmark with non-homogeneous hazy and
  haze-free images.
\newblock In {\em CVPRW}, 2020.

\bibitem{berman2016non}
Dana Berman, Shai Avidan, et~al.
\newblock Non-local image dehazing.
\newblock In {\em CVPR}, 2016.

\bibitem{bijelic2020seeing}
Mario Bijelic, Tobias Gruber, Fahim Mannan, Florian Kraus, Werner Ritter, Klaus
  Dietmayer, and Felix Heide.
\newblock Seeing through fog without seeing fog: Deep multimodal sensor fusion
  in unseen adverse weather.
\newblock In {\em CVPR}, 2020.

\bibitem{cai2016dehazenet}
Bolun Cai, Xiangmin Xu, Kui Jia, Chunmei Qing, and Dacheng Tao.
\newblock Dehazenet: An end-to-end system for single image haze removal.
\newblock {\em TIP}, 2016.

\bibitem{cao2021video}
Jiezhang Cao, Yawei Li, Kai Zhang, and Luc Van~Gool.
\newblock Video super-resolution transformer.
\newblock {\em arXiv preprint arXiv:2106.06847}, 2021.

\bibitem{chan2021basicvsr}
Kelvin~CK Chan, Xintao Wang, Ke Yu, Chao Dong, and Chen~Change Loy.
\newblock Basicvsr: The search for essential components in video
  super-resolution and beyond.
\newblock In {\em CVPR}, 2021.

\bibitem{chan2021basicvsr++}
Kelvin~CK Chan, Shangchen Zhou, Xiangyu Xu, and Chen~Change Loy.
\newblock Basicvsr++: Improving video super-resolution with enhanced
  propagation and alignment.
\newblock In {\em CVPR}, 2022.

\bibitem{chen2016robust}
Chen Chen, Minh~N Do, and Jue Wang.
\newblock Robust image and video dehazing with visual artifact suppression via
  gradient residual minimization.
\newblock In {\em ECCV}, 2016.

\bibitem{chen2019gated}
Dongdong Chen, Mingming He, Qingnan Fan, Jing Liao, Liheng Zhang, Dongdong Hou,
  Lu Yuan, and Gang Hua.
\newblock Gated context aggregation network for image dehazing and deraining.
\newblock In {\em WACV}, 2019.

\bibitem{cordts2016cityscapes}
Marius Cordts, Mohamed Omran, Sebastian Ramos, Timo Rehfeld, Markus Enzweiler,
  Rodrigo Benenson, Uwe Franke, Stefan Roth, and Bernt Schiele.
\newblock The cityscapes dataset for semantic urban scene understanding.
\newblock In {\em CVPR}, 2016.

\bibitem{dai2017deformable}
Jifeng Dai, Haozhi Qi, Yuwen Xiong, Yi Li, Guodong Zhang, Han Hu, and Yichen
  Wei.
\newblock Deformable convolutional networks.
\newblock In {\em ICCV}, 2017.

\bibitem{deng2019deep}
Zijun Deng, Lei Zhu, Xiaowei Hu, Chi-Wing Fu, Xuemiao Xu, Qing Zhang, Jing Qin,
  and Pheng-Ann Heng.
\newblock Deep multi-model fusion for single-image dehazing.
\newblock In {\em ICCV}, 2019.

\bibitem{dong2020multi}
Hang Dong, Jinshan Pan, Lei Xiang, Zhe Hu, Xinyi Zhang, Fei Wang, and
  Ming-Hsuan Yang.
\newblock Multi-scale boosted dehazing network with dense feature fusion.
\newblock In {\em CVPR}, 2020.

\bibitem{dosovitskiy2020image}
Alexey Dosovitskiy, Lucas Beyer, Alexander Kolesnikov, Dirk Weissenborn,
  Xiaohua Zhai, Thomas Unterthiner, Mostafa Dehghani, Matthias Minderer, Georg
  Heigold, Sylvain Gelly, et~al.
\newblock An image is worth 16x16 words: Transformers for image recognition at
  scale.
\newblock In {\em ICLR}, 2020.

\bibitem{fu2018deep}
Huan Fu, Mingming Gong, Chaohui Wang, Kayhan Batmanghelich, and Dacheng Tao.
\newblock Deep ordinal regression network for monocular depth estimation.
\newblock In {\em CVPR}, 2018.

\bibitem{guizilini20203d}
Vitor Guizilini, Rares Ambrus, Sudeep Pillai, Allan Raventos, and Adrien
  Gaidon.
\newblock 3d packing for self-supervised monocular depth estimation.
\newblock In {\em CVPR}, 2020.

\bibitem{guo2022image}
Chun-Le Guo, Qixin Yan, Saeed Anwar, Runmin Cong, Wenqi Ren, and Chongyi Li.
\newblock Image dehazing transformer with transmission-aware 3d position
  embedding.
\newblock In {\em CVPR}, 2022.

\bibitem{he2010single}
Kaiming He, Jian Sun, and Xiaoou Tang.
\newblock Single image haze removal using dark channel prior.
\newblock {\em TPAMI}, 2010.

\bibitem{huang2022neural}
Cong Huang, Jiahao Li, Bin Li, Dong Liu, and Yan Lu.
\newblock Neural compression-based feature learning for video restoration.
\newblock In {\em CVPR}, 2022.

\bibitem{huang2022monodtr}
Kuan-Chih Huang, Tsung-Han Wu, Hung-Ting Su, and Winston~H Hsu.
\newblock Monodtr: Monocular 3d object detection with depth-aware transformer.
\newblock In {\em CVPR}, 2022.

\bibitem{kim2020video}
Dahun Kim, Sanghyun Woo, Joon-Young Lee, and In~So Kweon.
\newblock Video panoptic segmentation.
\newblock In {\em CVPR}, 2020.

\bibitem{kim2013optimized}
Jin-Hwan Kim, Won-Dong Jang, Jae-Young Sim, and Chang-Su Kim.
\newblock Optimized contrast enhancement for real-time image and video
  dehazing.
\newblock {\em Journal of Visual Communication and Image Representation}, 2013.

\bibitem{kim2018spatio}
Tae~Hyun Kim, Mehdi~SM Sajjadi, Michael Hirsch, and Bernhard Scholkopf.
\newblock Spatio-temporal transformer network for video restoration.
\newblock In {\em ECCV}, 2018.

\bibitem{kopf2021robust}
Johannes Kopf, Xuejian Rong, and Jia-Bin Huang.
\newblock Robust consistent video depth estimation.
\newblock In {\em CVPR}, 2021.

\bibitem{lai2018learning}
Wei-Sheng Lai, Jia-Bin Huang, Oliver Wang, Eli Shechtman, Ersin Yumer, and
  Ming-Hsuan Yang.
\newblock Learning blind video temporal consistency.
\newblock In {\em ECCV}, 2018.

\bibitem{li2017aod}
Boyi Li, Xiulian Peng, Zhangyang Wang, Jizheng Xu, and Dan Feng.
\newblock Aod-net: All-in-one dehazing network.
\newblock In {\em ICCV}, 2017.

\bibitem{li2018end}
Boyi Li, Xiulian Peng, Zhangyang Wang, Jizheng Xu, and Dan Feng.
\newblock End-to-end united video dehazing and detection.
\newblock In {\em AAAI}, 2018.

\bibitem{li2018benchmarking}
Boyi Li, Wenqi Ren, Dengpan Fu, Dacheng Tao, Dan Feng, Wenjun Zeng, and
  Zhangyang Wang.
\newblock Benchmarking single-image dehazing and beyond.
\newblock {\em TIP}, 2018.

\bibitem{li2018single}
Runde Li, Jinshan Pan, Zechao Li, and Jinhui Tang.
\newblock Single image dehazing via conditional generative adversarial network.
\newblock In {\em CVPR}, 2018.

\bibitem{li2015simultaneous}
Zhuwen Li, Ping Tan, Robby~T Tan, Danping Zou, Steven Zhiying~Zhou, and
  Loong-Fah Cheong.
\newblock Simultaneous video defogging and stereo reconstruction.
\newblock In {\em CVPR}, 2015.

\bibitem{liang2022recurrent}
Jingyun Liang, Yuchen Fan, Xiaoyu Xiang, Rakesh Ranjan, Eddy Ilg, Simon Green,
  Jiezhang Cao, Kai Zhang, Radu Timofte, and Luc Van~Gool.
\newblock Recurrent video restoration transformer with guided deformable
  attention.
\newblock In {\em NeurlPS}, 2022.

\bibitem{lin2022flow}
Jing Lin, Yuanhao Cai, Xiaowan Hu, Haoqian Wang, Youliang Yan, Xueyi Zou,
  Henghui Ding, Yulun Zhang, Radu Timofte, and Luc Van~Gool.
\newblock Flow-guided sparse transformer for video deblurring.
\newblock In {\em ICML}, 2022.

\bibitem{liu2019griddehazenet}
Xiaohong Liu, Yongrui Ma, Zhihao Shi, and Jun Chen.
\newblock Griddehazenet: Attention-based multi-scale network for image
  dehazing.
\newblock In {\em ICCV}, 2019.

\bibitem{liu2022phase}
Ye Liu, Liang Wan, Huazhu Fu, Jing Qin, and Lei Zhu.
\newblock Phase-based memory network for video dehazing.
\newblock In {\em ACM MM}, 2022.

\bibitem{liu2021synthetic}
Ye Liu, Lei Zhu, Shunda Pei, Huazhu Fu, Jing Qin, Qing Zhang, Liang Wan, and
  Wei Feng.
\newblock From synthetic to real: Image dehazing collaborating with unlabeled
  real data.
\newblock In {\em ACM MM}, 2021.

\bibitem{liu2021swin}
Ze Liu, Yutong Lin, Yue Cao, Han Hu, Yixuan Wei, Zheng Zhang, Stephen Lin, and
  Baining Guo.
\newblock Swin transformer: Hierarchical vision transformer using shifted
  windows.
\newblock In {\em ICCV}, 2021.

\bibitem{liu2022convnet}
Zhuang Liu, Hanzi Mao, Chao-Yuan Wu, Christoph Feichtenhofer, Trevor Darrell,
  and Saining Xie.
\newblock A convnet for the 2020s.
\newblock In {\em CVPR}, 2022.

\bibitem{mccartney1976optics}
Earl~J McCartney.
\newblock Optics of the atmosphere: scattering by molecules and particles.
\newblock {\em New York}, 1976.

\bibitem{nah2019ntire}
Seungjun Nah, Sungyong Baik, Seokil Hong, Gyeongsik Moon, Sanghyun Son, Radu
  Timofte, and Kyoung Mu~Lee.
\newblock Ntire 2019 challenge on video deblurring and super-resolution:
  Dataset and study.
\newblock In {\em CVPRW}, 2019.

\bibitem{oh2019video}
Seoung~Wug Oh, Joon-Young Lee, Ning Xu, and Seon~Joo Kim.
\newblock Video object segmentation using space-time memory networks.
\newblock In {\em ICCV}, 2019.

\bibitem{pont20172017}
Jordi Pont-Tuset, Federico Perazzi, Sergi Caelles, Pablo Arbel{\'a}ez, Alex
  Sorkine-Hornung, and Luc Van~Gool.
\newblock The 2017 davis challenge on video object segmentation.
\newblock {\em arXiv:1704.00675}, 2017.

\bibitem{qin2020ffa}
Xu Qin, Zhilin Wang, Yuanchao Bai, Xiaodong Xie, and Huizhu Jia.
\newblock Ffa-net: Feature fusion attention network for single image dehazing.
\newblock In {\em AAAI}, 2020.

\bibitem{qu2019enhanced}
Yanyun Qu, Yizi Chen, Jingying Huang, and Yuan Xie.
\newblock Enhanced pix2pix dehazing network.
\newblock In {\em CVPR}, 2019.

\bibitem{ranjan2017optical}
Anurag Ranjan and Michael~J Black.
\newblock Optical flow estimation using a spatial pyramid network.
\newblock In {\em CVPR}, 2017.

\bibitem{ren2016single}
Wenqi Ren, Si Liu, Hua Zhang, Jinshan Pan, Xiaochun Cao, and Ming-Hsuan Yang.
\newblock Single image dehazing via multi-scale convolutional neural networks.
\newblock In {\em ECCV}, 2016.

\bibitem{ren2018deep}
Wenqi Ren, Jingang Zhang, Xiangyu Xu, Lin Ma, Xiaochun Cao, Gaofeng Meng, and
  Wei Liu.
\newblock Deep video dehazing with semantic segmentation.
\newblock {\em TIP}, 2018.

\bibitem{sakaridis2018semantic}
Christos Sakaridis, Dengxin Dai, and Luc Van~Gool.
\newblock Semantic foggy scene understanding with synthetic data.
\newblock {\em IJCV}, 2018.

\bibitem{shi2016real}
Wenzhe Shi, Jose Caballero, Ferenc Husz{\'a}r, Johannes Totz, Andrew~P Aitken,
  Rob Bishop, Daniel Rueckert, and Zehan Wang.
\newblock Real-time single image and video super-resolution using an efficient
  sub-pixel convolutional neural network.
\newblock In {\em CVPR}, 2016.

\bibitem{song2023vision}
Yuda Song, Zhuqing He, Hui Qian, and Xin Du.
\newblock Vision transformers for single image dehazing.
\newblock {\em TIP}, 2023.

\bibitem{tassano2020fastdvdnet}
Matias Tassano, Julie Delon, and Thomas Veit.
\newblock Fastdvdnet: Towards real-time deep video denoising without flow
  estimation.
\newblock In {\em CVPR}, 2020.

\bibitem{tian2020tdan}
Yapeng Tian, Yulun Zhang, Yun Fu, and Chenliang Xu.
\newblock Tdan: Temporally-deformable alignment network for video
  super-resolution.
\newblock In {\em CVPR}, 2020.

\bibitem{vaswani2017attention}
Ashish Vaswani, Noam Shazeer, Niki Parmar, Jakob Uszkoreit, Llion Jones,
  Aidan~N Gomez, {\L}ukasz Kaiser, and Illia Polosukhin.
\newblock Attention is all you need.
\newblock {\em NeurIPS}, 2017.

\bibitem{wang2019edvr}
Xintao Wang, Kelvin~CK Chan, Ke Yu, Chao Dong, and Chen Change~Loy.
\newblock Edvr: Video restoration with enhanced deformable convolutional
  networks.
\newblock In {\em CVPRW}, 2019.

\bibitem{wen2020ua}
Longyin Wen, Dawei Du, Zhaowei Cai, Zhen Lei, Ming-Ching Chang, Honggang Qi,
  Jongwoo Lim, Ming-Hsuan Yang, and Siwei Lyu.
\newblock Ua-detrac: A new benchmark and protocol for multi-object detection
  and tracking.
\newblock {\em Computer Vision and Image Understanding}, 2020.

\bibitem{wu2021contrastive}
Haiyan Wu, Yanyun Qu, Shaohui Lin, Jian Zhou, Ruizhi Qiao, Zhizhong Zhang, Yuan
  Xie, and Lizhuang Ma.
\newblock Contrastive learning for compact single image dehazing.
\newblock In {\em CVPR}, 2021.

\bibitem{xia2022vision}
Zhuofan Xia, Xuran Pan, Shiji Song, Li~Erran Li, and Gao Huang.
\newblock Vision transformer with deformable attention.
\newblock In {\em CVPR}, 2022.

\bibitem{yang2019frame}
Wenhan Yang, Jiaying Liu, and Jiashi Feng.
\newblock Frame-consistent recurrent video deraining with dual-level flow.
\newblock In {\em CVPR}, 2019.

\bibitem{zhang2018densely}
He Zhang and Vishal~M Patel.
\newblock Densely connected pyramid dehazing network.
\newblock In {\em CVPR}, 2018.

\bibitem{zhang2021learning}
Xinyi Zhang, Hang Dong, Jinshan Pan, Chao Zhu, Ying Tai, Chengjie Wang, Jilin
  Li, Feiyue Huang, and Fei Wang.
\newblock Learning to restore hazy video: A new real-world dataset and a new
  method.
\newblock In {\em CVPR}, 2021.

\bibitem{zhao2017pyramid}
Hengshuang Zhao, Jianping Shi, Xiaojuan Qi, Xiaogang Wang, and Jiaya Jia.
\newblock Pyramid scene parsing network.
\newblock In {\em CVPR}, 2017.

\bibitem{zheng2021ultra}
Zhuoran Zheng, Wenqi Ren, Xiaochun Cao, Xiaobin Hu, Tao Wang, Fenglong Song,
  and Xiuyi Jia.
\newblock Ultra-high-definition image dehazing via multi-guided bilateral
  learning.
\newblock In {\em CVPR}, 2021.

\bibitem{zhu2021detection}
Pengfei Zhu, Longyin Wen, Dawei Du, Xiao Bian, Heng Fan, Qinghua Hu, and Haibin
  Ling.
\newblock Detection and tracking meet drones challenge.
\newblock {\em TPAMI}, 2021.

\end{thebibliography}
}

\end{document}